%% file: main.tex
\documentclass{article}

\PassOptionsToPackage{numbers, compress}{natbib}
\usepackage[final]{neurips_2019}

\usepackage[utf8]{inputenc} 
\usepackage[T1]{fontenc}    
\usepackage{hyperref}       
\usepackage{url}            
\usepackage{booktabs}       
\usepackage{amsfonts}       
\usepackage{nicefrac}       
\usepackage{microtype}      
\usepackage{placeins}

\usepackage{bm}
\usepackage{xspace}
\RequirePackage{algorithm}
\RequirePackage{algorithmic}
\usepackage{wrapfig}
\usepackage{subcaption} 
\usepackage{multirow}
\usepackage{adjustbox}
\usepackage{amsmath}
\usepackage{graphicx}
\usepackage{afterpage}

\usepackage{todonotes}

\title{A Model to Search for Synthesizable Molecules}

\date{}

\newcommand{\selectro}{\textsc{Molecule Chef}\xspace}

\include{tex/notation}

\newcommand{\affil}[1]{\ifcase#1\or$\dagger$\or$\ddagger$\or$\mathparagraph$\or$\mathsection$\or$\sharp$\or$\clubsuit$\fi}

\author{%
  John Bradshaw \\
 University of Cambridge \\
   MPI for Intelligent Systems\\
  \texttt{jab255@cam.ac.uk} \\
   \And
  Brooks Paige \\
  University of Cambridge \\
  The Alan Turing Institute  \\
  \texttt{bpaige@turing.ac.uk} \\
   \And
   Matt J. Kusner \\
University College London \\
  The Alan Turing Institute  \\
   \texttt{m.kusner@ucl.ac.uk} \\
   \And
    Marwin H. S. Segler\ \\
  BenevolentAI  \\
  Westf\"alische Wilhelms-Universit\"at M\"unster\\
   \texttt{marwin.segler@benevolent.ai} \\
   \And
    Jos\'e Miguel Hern\'andez-Lobato \\
   University of Cambridge \\
   The Alan Turing Institute\\
   Microsoft Research Cambridge\\
   \texttt{jmh233@cam.ac.uk} \\
}

\begin{document}

\maketitle
\begin{abstract}
\input{tex/abstract}

\end{abstract}
\section{Introduction}
\input{tex/intro}

\section{Background}
\label{sect:background}
\input{tex/background}

\section{Model}
\input{tex/model}

\section{Evaluation}
\input{tex/evaluation}

\section{Discussion}
\input{tex/conclusion}

\subsection*{Acknowledgements}
This work was supported by The Alan Turing Institute under the EPSRC grant EP/N510129/1. 
JB also acknowledges support from an EPSRC studentship.

\bibliography{refs}
\bibliographystyle{plainnat}

\newpage
\appendix
\section{Appendix}
\input{tex/appendix}

\end{document}

%% file: tex/notation.tex
\newcommand{\allMolecules}{\mathcal{G}}
\newcommand{\reactantsPool}{\mathcal{R}}
\newcommand{\graphRepresentations}{\mathbf{m}_g}
\newcommand{\graph}{g}

%% file: tex/abstract.tex
%
\noindent%
Deep generative models are able to suggest new organic molecules by generating strings, trees, and graphs representing their structure. 
While such models allow one to generate molecules with desirable properties, they give no guarantees that the molecules can actually be synthesized in practice. 
We propose a new molecule generation model, mirroring a more realistic real-world process, where (a) reactants are selected, and (b) combined to form more complex molecules. 
More specifically, our generative model proposes a bag of initial reactants (selected from a pool of commercially-available molecules) and uses a reaction model to predict how they react together to generate new molecules. 
We first show that the model can generate diverse, valid and unique molecules due to the useful inductive biases of modeling reactions. 
Furthermore, our model allows chemists to interrogate not only the properties of the generated molecules but also the feasibility of the synthesis routes.
We conclude by using our model to solve retrosynthesis problems, predicting a set of reactants that can produce a target product.

%% file: tex/intro.tex
The ability of machine learning to generate structured objects has progressed dramatically in the last few years. 
One particularly successful example of this is the flurry of developments devoted to generating small molecules
\citep{Gomez-Bombarelli2018-ex,segler2017generating,Kusner2017-ry,dai2018syntax,Simonovsky2018-md,De_Cao2018-sq,Jin2018-aa,Liu2018-ha,you2018graph,samanta2019nevae, assouel2018defactor}. 
These models have been shown to be extremely effective at finding molecules with desirable properties: drug-like molecules \citep{Gomez-Bombarelli2018-ex},
biological target activity molecules \citep{segler2017generating}, and soluble molecules \citep{De_Cao2018-sq}.

However, these improvements in molecule discovery come at a cost: these methods do not describe \emph{how to synthesize such molecules}, a prerequisite for experimental testing. 
Traditionally, in computer-aided molecular design, this has been addressed by virtual screening \citep{shoichet2004virtual}, where molecule data sets $|D|\approx 10^8$, are first generated via
the expensive combinatorial enumeration of molecular fragments stitched together using hand-crafted bonding rules, and then are scored in an $\mathcal{O}(|D|)$ step.

\begin{figure*}[t!]
 \centering
  \includegraphics[width=\columnwidth]{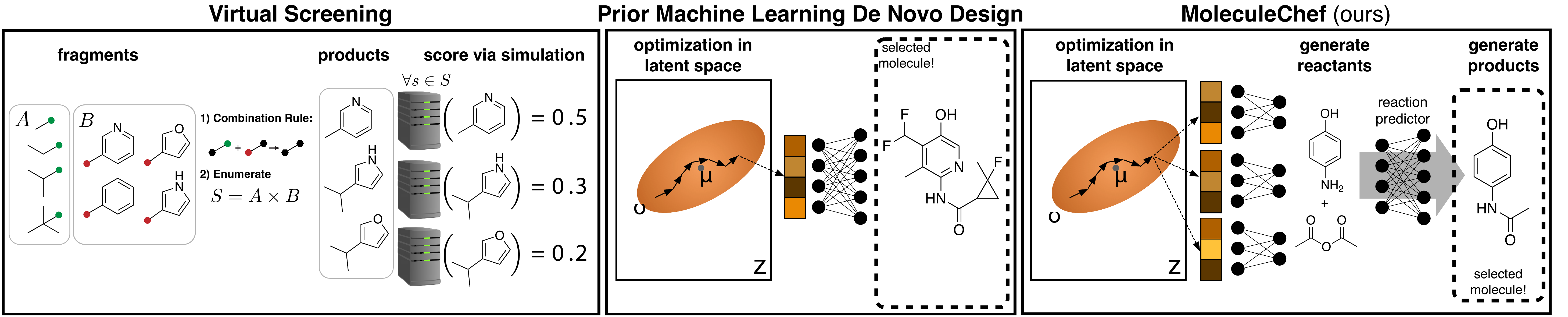}
  \caption{
    An overview of approaches used to find molecules with desirable properties.
    \emph{Left}: Virtual screening \citep{shoichet2004virtual} aims to find novel molecules by the (computationally expensive) enumeration over all possible combinations of fragments.
    \emph{Center}: More recent ML approaches, eg \citep{Gomez-Bombarelli2018-ex}, aim to find useful, novel molecules by optimizing in a continuous latent space; however, there are no clues to whether (and how)
    these molecules can be synthesized.
  \emph{Right}: We approach the generation of molecules through a multistage process mirroring how complex molecules are created in practice, while maintaining a continuous latent space to use for optimization.
Our model, \selectro, first finds suitable reactants which then react together to create a final molecule.}
  \label{fig:modelOverview}
\end{figure*}

In this paper we propose a generative model for molecules (shown in Figure \ref{fig:modelOverview}) that describes how to make such molecules from a set of commonly-available reactants. 
Our model first generates a set of reactant molecules, and second maps them to a predicted product molecule via a reaction prediction model. 
It allows one to simultaneously search for better molecules and describe how such molecules can be made. By closely mimicking the real-world process of designing new molecules, we show that our model:
1. Is able to generate a wide range of molecules not seen in the training data; 2. Addresses practical synthesis concerns such as reaction stability and toxicity; 
and 3. Allows us to propose new \emph{reactants} for given target molecules that may be more practical to manage. 

%% file: tex/background.tex

We start with an overview of traditional computational techniques to discover novel molecules with desirable properties. 
We then review recent work in machine learning (ML) that seeks to improve parts of this process.
We then identify aspects of molecule discovery we believe deserve much more attention from the ML community.
We end by laying out our contributions to address these concerns.

\subsection{Virtual Screening}
To discover new molecules with certain properties, one popular technique is \emph{virtual screening} (VS) \citep{shoichet2004virtual,hu2011leap,pyzer2015high, chevillard2015scubidoo,nicolaou2016proximal}.
VS works by 
(a) enumerating all combinations of a set of building-block molecules (which are combined via virtual chemical bonding rules),
(b) for each molecule, calculating the desired properties via simulations or prediction models, 
(c) filtering the most interesting molecules to synthesize in the lab. While VS is general, 
it has the important downside that the generation process is not targeted: VS needs to get lucky to find molecules with desirable properties, it does not search for them.
Given that the number of possible drug-like compounds is estimated to be $\in [10^{23},10^{100}]$ \citep{van2019virtual}, the chemical space usually screened 
in VS $\in [10^7, 10^{10}]$ is tiny. Searching in combinatorial fragment spaces has been proposed, but is limited to simpler similarity queries \cite{rarey2001similarity}.

\subsection{The Molecular Search Problem}
\label{sect:mol-search-problem}

To address these downsides, one idea is to replace this full enumeration with a search algorithm; an idea called \emph{de novo-design} (DND) \citep{schneider2016novo}. 
Instead of generating a large set of molecules with small variations, DND searches for molecules with particular properties, recomputes them for the newfound molecules, and searches again. 
We call this \textbf{the molecular search problem}. 
Early work on the molecular search problem used genetic algorithms, ant-colony optimization, or other discrete search techniques to make local changes to molecules \citep{hartenfeller2011enabling}. 
While more directed than library-generation, these approaches still explored locally, limiting the diversity of discovered molecules. 

The first work to apply current ML techniques to this problem was \citet{Gomez-Bombarelli2018-ex} (in a late 2016 preprint). 
Their idea was to search by learning a mapping from molecular space to continuous space and back. With this mapping 
it is possible to leverage well-studied optimization techniques to do search: local search can be done via gradient descent and global search via Bayesian optimization \citep{snoek2012practical,gardner2014bayesian}.
For such a mapping, the authors chose to represent molecules as SMILES strings \citep{weininger1988smiles} and leverage advances in generative models for text \citep{bowman2015generating} to learn a character variational autoencoder (CVAE) \citep{kingma2013auto}. 
Shortly after this work, in an early 2017 preprint, \citet{segler2017generating} trained recurrent neural networks (RNNs) to take properties as input and output SMILES strings with these properties, with molecular search done using reinforcement learning (RL).

\paragraph{In Search of Molecular Validity.} 
However, the SMILES string representation is very brittle: 
if individual characters are changed or swapped, it may no longer represent any molecule (called an \emph{invalid} molecule). Thus, the CVAE 
often produced invalid molecules (in one experiment, \citet{Kusner2017-ry} sampling from the continuous space, produced valid molecules only $0.7\%$ of the time). To address this
 validity problem, recent works have proposed using alternative molecular representations such as parse trees \citep{Kusner2017-ry,dai2018syntax} 
 or graphs \citep{Simonovsky2018-md,De_Cao2018-sq,Li2018-zg,Jin2018-aa,Liu2018-ha,you2018graph,jin2018learning,pmlr-v97-kajino19a,samanta2019nevae}, where
 some of the more recent among these enforce or strongly encourage
 validity  \citep{Jin2018-aa,Liu2018-ha,you2018graph,pmlr-v97-kajino19a,samanta2019nevae}.
 In parallel, there has been work based on RL that has aimed to learn a validity function during training directly  \citep{guimaraes2017objective,janz2017learning}.

\subsection{The Molecular Recipe Problem}
Crucially, all of the works in the previous section solving \textbf{the molecular search problem} focus purely on optimizing molecules towards desirable properties.
These works, in addressing the downsides of VS, removed a benefit of it: knowledge of the synthesis pathway of each molecule. Without this we do not know   
\emph{how practical it is to make ML-generated molecules}. 

To address this concern is to address \textbf{the molecular recipe problem}: 
what molecules are we able to make, given a set of readily-available starting molecules?  
So far, this problem has been addressed independently of the molecular search problem through synthesis planning (SP) \citep{segler2018planning}. SP works by recursively deconstructing a molecule.
This deconstruction is done via (reversed) 
\emph{reaction predictors}: models that predict how reactant molecules produce a product molecule. 
More recently, novel ML models have been designed for reaction prediction \citep{wei2016neural,segler2017neural,Jin2017-hh,schwaller2017found, bradshaw2018generative, schwaller2018molecular}.

\subsection{This Work}
In this paper, we propose to address both \textbf{the molecular search problem} and \textbf{the molecular recipe problem} jointly. To do so, we propose a generative model over molecules using the following map: 
First, a mapping from continuous space to a set of known, reliable, easy-to-obtain reactant molecules. 
Second a mapping from this set of reactant molecules to a final product molecule, based on a reaction prediction model \citep{wei2016neural,segler2017neural,Jin2017-hh,schwaller2018molecular,bradshaw2018generative}.
Thus our generative model not only generates molecules, but also \emph{ a synthesis route using available reactants}.
This addresses the molecular recipe problem, and also the molecular search problem, as the learned continuous space can also be used for search. 
Compared to previous work, in this work we are searching for new molecules through virtual chemical reactions, more directly simulating how molecules are actually discovered in the lab.

Concretely, we argue that our model, which we shall introduce in the next section, has several advantages over the current deep generative models of molecules reviewed previously:
\begin{description}
  \item[Better extrapolation properties]
   Generating molecules through graph editing operations, representing reactions, we hope gives us strong inductive biases for extrapolating well.
  \item[Validity of generated molecules] 
  Naive generation of molecular SMILES strings or graphs can lead to molecules that are invalid. 
  Although the syntactic validity can be fixed by using masking \citep{Kusner2017-ry, Liu2018-ha}, the molecules generated can often still be semantically invalid. 
   By generating molecules from chemically stable reactants by means of reactions, our model proposes more semantically valid molecules.
  \item[Provide synthesis routes] Proposed molecules from other methods can often not be evaluated in practice, as chemists do not know how to synthesize them. 
  As a byproduct of our model we suggest synthetic routes, which could have a useful, practical value.
\end{description}

%% file: tex/model.tex

In this section we describe our model\footnote{Further details can also be found in
our appendix and code is available at \url{https://github.com/john-bradshaw/molecule-chef}}.
We define the set of all possible valid molecular graphs as $\allMolecules$, with an individual graph 
$\graph \in \allMolecules$ representing the atoms of a molecule as its nodes, and the type of bonds between these atoms (we consider single, double and triple bonds) as its edge types.
The set of common reactant molecules, easily procurable by a chemist, which we want to act as building blocks for any final molecule is a subset of this, $\reactantsPool \subset \allMolecules$.

As discussed in the previous section (and shown in Figure \ref{fig:modelOverview}) our generative model for molecules consists of the composition of two parts: 
(1) a decoder from a continuous latent space, $\mathbf{z} \in \mathbb{R}^m$, to a bag (ie multiset\footnote{Note how we allow molecules to be present multiple times as reactants in our reaction,
although practically many reactions only have one instance of a particular reactant.}) of easily procurable reactants, $\mathbf{x} \subset \reactantsPool$;
(2) a reaction predictor model that transforms this bag of molecules into a multiset of product molecules $\mathbf{y} \subset \allMolecules$. 

The benefit of this approach is that for step (2) we can pick from several existing reaction predictor models, including recently proposed methods that have used ML techniques
\citep{kayala2011learning,segler2017neural,schwaller2018molecular,bradshaw2018generative,coley2019graph}.
In this work we use the Molecular Transformer (MT) of \citet{schwaller2018molecular}, as it has recently been shown to provide state-of-the-art performance in this task \citep[Table 4]{schwaller2018molecular}. 

This leaves us with the task of (1), learning a way to decode to (and encode from) a bag of reactants, using a parameterized encoder $q(\mathbf z|\mathbf x )$ and decoder $p(\mathbf x|\mathbf z)$. 
We call this co-occurrence model \selectro, and by moving around in the latent space we can \emph{select} using \selectro different ``bags of reactants''.

Again there are several viable options of how to learn \selectro.
For instance one could choose to use a VAE for this task \citep{kingma2013auto, rezende2014stochastic}.
However, when paired with a complex decoder these models are often difficult to train \citep{bowman2015generating, alemi2018fixing}, such that much of the previous work for generating graphs has  has tuned down the KL regularization term in these models \citep{Liu2018-ha, Kusner2017-ry}.
We therefore instead propose using the WAE objective \citep{tolstikhin2017wasserstein}, which involves minimizing
\begin{align*}
	L = & \mathbb{E}_{\mathbf{x}  \sim \mathcal{D}} \mathbb{E}_{q(\mathbf z |\mathbf x)} \left[ c(\mathbf{x}, p(\mathbf{x}|\mathbf{z})) \right] 
	  + \lambda D \left( \mathbb{E}_{\mathbf{x}  \sim \mathcal{D}} \left[q(\mathbf z|\mathbf x)\right], p(\mathbf{z}) \right)
\end{align*}
where $c$ is a cost function, that enforces the reconstructed bag to be similar to the encoded one. $D$ is a divergence measure, which is weighted in relative importance by $\lambda$, that forces the marginalised distribution of all encodings to match the prior on the latent space.
 Following \citet{tolstikhin2017wasserstein} we use the maximum mean discrepancy (MMD) divergence measure, with $\lambda=10$ and a standard normal prior over the latents.
 We choose $c$ so that this first term matches the reconstruction term we would obtain in a VAE, i.e.\ with
 $c(\mathbf{x}, \mathbf{z}) = - \log p(\mathbf{x} | \mathbf{z})$.
This means that the objective only differs from a VAE in the second, regularisation term, 
such that we are not trying to match each encoding to the prior but instead the marginalised distribution over all datapoints.
Empirically, we find that this trains well and does not suffer from the same local optimum issues as the VAE.

\subsection{Encoder and Decoder}

We can now begin describing the structure of our encoder and decoder.
In these functions it is often convenient to work with $n$-dimensional vector embeddings of graphs, $\graphRepresentations \in \mathbb{R}^n$.
Again we are faced with a series of possible alternative ways to compute these embeddings.
For instance, we could ignore the structure of the molecular graph and learn a distinct embedding for each molecule, or use fixed molecular fingerprints, such as Morgan Fingerprints \citep{morgan1965generation}. 
We instead choose to use deep graph neural networks \citep{merkwirth2005automatic,duvenaud2015convolutional,battaglia2018relational} that can produce graph-isomorphic representations.

Deep graph neural networks have been shown to perform well on a variety of tasks involving small organic molecules,
 and their advantages compared to the previously mentioned alternative approaches are that (1) they take the structure of the graph into account and (2) they can learn which characteristics are important when forming higher-level representations.
In particular in this work we use 4 layer Gated Graph Neural Networks (GGNN) \citep{li2015gated}. 
These compute higher-level representations for each node.
These node-level representations in turn can be combined by a weighted sum, to form a graph-level representation invariant to the order of the nodes, 
in an operation referred to as an aggregation transformation \citep[\S3]{Johnson2017-pd}.

\begin{figure*}[t]
\centering
  \includegraphics[trim={.3em .8em .3em .3em},clip,width=0.85\textwidth]{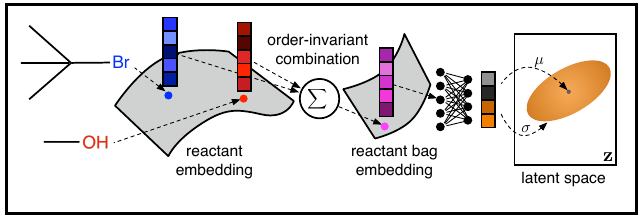}
  \caption{The encoder of \selectro. 
  This maps from a multiset of reactants to a distribution over latent space.
There are three main steps: (1) the reactants molecules are embedded into a continuous space by using GGNNs \citep{li2015gated} to form molecule embeddings;
(2) the molecule embeddings in the multiset are summed to form one order-invariant embedding for the whole multiset; 
(3) this is then used as input to a neural network which parameterizes a Gaussian distribution over $\mathbf{z}$.}
  \label{fig:encoder}
\end{figure*}

\paragraph{Encoder}

The structure of \selectro's encoder, $q(\mathbf z|\mathbf x )$, is shown in Figure \ref{fig:encoder}.
For the $i$th data point the encoder has as input the multiset of reactants $\mathbf{x_i} = \{x^i_1, x^i_2, \cdots \}$.
It first computes the representation of each individual reactant molecule graph using the GGNN, before summing these representations to get a representation that is invariant to the order of the multiset. 
A feed forward network is then used to parameterize the mean and variance of a Gaussian distribution over $\mathbf{z}$.

\paragraph{Decoder}

\begin{figure*}[t]
\centering
  \includegraphics[trim={.3em .3em .3em .3em},clip,width=0.85\textwidth]{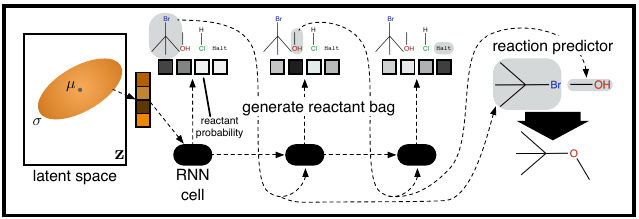}
  \caption{The decoder of \selectro. The decoder generates the multiset of reactants in sequence through calls to a RNN.
  At each step the model picks either one reactant from the pool or to halt, finishing the sequence.
The latent vector, $\mathbf z$, is used to parameterize the initial hidden layer of the RNN.
Reactants that are selected are fed back into the RNN on the next step.
The reactant bag formed is later fed through a reaction predictor to form a final product.}
  \label{fig:decoder}
\end{figure*}

The decoder, $p(\mathbf x|\mathbf z)$, (Figure~\ref{fig:decoder}) maps from the latent space to a multiset of reactant molecules. 
These reactants are typically small molecules, which means we could fit a deep generative model which produces them from scratch.
However, to better mimic the process of selecting reactant molecules from an easily obtainable set, we instead restrict the output 
of the decoder to pick the molecules from a fixed set of reactant molecules, $\reactantsPool$.

This happens in a sequential process using a recurrent neural network (RNN), with the full process described in Algorithm \ref{alg:decoder}.
The latent vector, $\mathbf z$ is used to parametrize the initial hidden layer of the RNN.
The selected reactants are fed back in as inputs to the RNN at the next generation stage.
Whilst training we randomly sample the ordering of the reactants, and use teacher forcing.

\begin{algorithm}[t]
  \small
  \caption{\selectro's Decoder}
  \label{alg:decoder}
  \begin{algorithmic}
    \REQUIRE $\mathbf z^i$ (latent space sample), GGNN (for embedding molecules), RNN (recurrent neural network), $\reactantsPool$ (set of easy-to-obtain reactant molecules), $\mathbf s$ (learnt ``halt'' embedding), $\mathbf A$ (learnt matrix that projects the size of the latent space to the size of RNN's hidden space)
    \STATE $\mathbf h_0 \leftarrow \mathbf A \mathbf z^i$ ; $ \mathbf m_0 \leftarrow  \mathbf 0$ 
    \COMMENT{Start symbol}
    \FOR{$t=1$ to $T_{\textrm{max}}$}
    \STATE $\mathbf h_t \leftarrow  \textrm{RNN}(\mathbf m_{t-1} , \mathbf h_{t-1})$ ; $\mathbf B  \leftarrow $ STACK([GGNN(g) for all g in $\reactantsPool$] + [$\mathbf s$])
    \STATE logits $\leftarrow \mathbf h_t \mathbf B^T$
    \STATE $ x_t \sim \textrm{softmax}(\textrm{logits})$
    \IF{$ x_t = \textrm{HALT}$}
    \STATE break
    \COMMENT{If the logit corresponding to the halt embedding is selected then we stop early}
    \ELSE
    \STATE $\mathbf m_t \leftarrow \textrm{GGNN}(x_t)$
    \ENDIF
    \ENDFOR
    \STATE return  $x_1, x_2, \cdots$
  \end{algorithmic}
\end{algorithm}

  

\subsection{Adding a predictive penalty loss to the latent space}
\label{sect:property-predictor}

As discussed in section \ref{sect:mol-search-problem} we are interested in using and evaluating our model's performance 
in the \textbf{molecular search problem}, that is using the learnt latent space to find new molecules with desirable properties.
In reality we would wish to measure some complex chemical property that can only be measured experimentally.
However, as a surrogate for this, following \citep{Gomez-Bombarelli2018-ex},  we optimize instead for the QED (Quantitative Estimate of Drug-likeness 
\citep{bickerton2012quantifying}) score of a molecule, $\mathbf w$, as a deterministic mapping from molecules to this score,
$\mathbf y \mapsto \mathbf w$, exists in RDKit \citep{rdkit}.

To this end, in a similar manner to \citet[\S 4.3]{Liu2018-ha} \& \citet[\S 3.3]{Jin2018-aa}, we can simultaneously train a 2 hidden layer property predictor NN for use in local optimization tasks.
This network tries to predict the QED property, $\mathbf w$, of the final product $\mathbf y$ from the latent encoding of the associated bag of reactants.
The use of this property predictor network for local optimization is described in Section \ref{sect:local-optimization}.

\FloatBarrier

%% file: tex/evaluation.tex

In this section we evaluate \selectro in 
(1) its ability to generate a diverse set of valid molecules; 
(2) how useful its learnt latent space is when optimizing product molecules for some property; 
and (3) whether by training a regressor back from product molecules to the latent space, \selectro can be used as part of a setup to perform retrosynthesis.

In order to train our model we need a dataset of reactant bags.
For this we use the USPTO dataset \citep{lowe2012extraction}, processed and cleaned up by \citet{Jin2017-hh}.
We filter out reagents, molecules that form context under which the reaction occurs but do not contribute atoms to the final products, by following the approach of \citet[\S3.1]{schwaller2017found}.

We wish to use as possible reactant molecules only popular molecules that a chemist would  have easy access to. 
To this end, we filter our training (using \citet{Jin2017-hh}'s split) dataset so that each reaction only contains reactants that occur at least 15 times across different reactions in the original larger training USPTO dataset.
This leaves us with a dataset of 34426 unique reactant bags for training the \selectro. 
In total there are 4344 unique reactants.
For training the baselines, we combine these 4344 unique reactants and the associated products from their different combinations, to form a training set for baselines, as even though \selectro has not seen the products during training, the reaction predictor has.

\subsection{Generation}

We begin by analyzing our model using the metrics favored by previous work\footnote{Note that we have extended the definition of these metrics to a
  bag (multiset) of products, given that our model can output multiple molecules for each reaction. However, when sampling 20000 times from the prior of our model,
  we generate single product bags 97\% of the time,
so that in practice most of the time we are using the same definition for these metrics as the previous work which always generated single molecules.}
 \citep{Jin2018-aa, Liu2018-ha, Li2018-zg, Kusner2017-ry}: validity, uniqueness and novelty. 
Validity is defined as requiring that at least one of the molecules in the bag of products can be parsed by RDKit.
For a bag of products to be unique we require it to have at least one valid molecule that the model has not generated before in any of the previously seen bags.
Finally, for computing novelty we require that the valid molecules not be present in the same training set we use for the baseline generative models.

\begin{table*}[h]
\begin{center}
    \caption{Table showing the validity, uniqueness, novelty and normalized quality  (all as \%, higher better) of the products/or molecules generated from decoding from 20k random samples from the prior $p(\mathbf z)$.
     Quality is the proportion of valid molecules that pass the quality filters proposed in \citet[\S3.3]{brown2018guacamol}, normalized such that the score on the training set is 100.
     FCD is the Fréchet ChemNet Distance \citep{doi:10.1021/acs.jcim.8b00234}, capturing a notion of distance between the generated valid molecules and the training dataset (lower better).
  The uniqueness and novelty figures are also conditioned on validity.
   MT stands for the Molecular Transformer \citep{schwaller2018molecular}. }
  \begin{tabular}{lrrrrr}
    \toprule
    Model Name & Validity & Uniqueness & Novelty & Quality & FCD \\
    \midrule
    \selectro + MT & 99.05	&	95.95	&	89.11	&	95.30	&	0.73 \\
  \midrule
  AAE \citep{kadurin2017cornucopia, polykovskiy2018molecular} & 85.86	&	98.54	&	93.37	&	94.89	&	1.12 \\
  CGVAE \citep{Liu2018-ha} & 100.00	&	93.51	&	95.88	&	44.45	&	11.73 \\
  CVAE \citep{Gomez-Bombarelli2018-ex} & 12.02	&	56.28	&	85.65	&	52.86	&	37.65 \\
  GVAE \citep{Kusner2017-ry} & 12.91	&	70.06	&	87.88	&	46.87	&	29.32 \\
  LSTM \citep{segler2017generating} & 91.18	&	93.42	&	74.03	&	100.12	&	0.43 \\
    \bottomrule
  \end{tabular}
\label{table:generation-properties}
    \end{center}
\end{table*}

In addition, we compute the Fréchet ChemNet Distance (FCD) \citep{doi:10.1021/acs.jcim.8b00234} between the valid molecules generated by each method and our baseline training set.
Finally in order to try to assess the \emph{quality} of the molecules generated we record the (train-normalized) proportion of valid molecules that pass the quality filters proposed
by \citet[\S3.3]{brown2018guacamol}; these filters aim to remove molecules that are \emph{``potentially unstable, reactive, laborious to
synthesize, or simply unpleasant to the eye of medicinal chemists''}.

For the baselines we consider the character VAE (CVAE) \citep{Gomez-Bombarelli2018-ex}, the grammar VAE (GVAE) \citep{Kusner2017-ry}, the AAE (adversarial autoencoder) \citep{kadurin2017cornucopia}, 
the constrained graph VAE (CGVAE) \citep{Liu2018-ha},
and a stacked LSTM generator with no latent space \citep{segler2017generating}.
Further details about the baselines can be found in the appendix.

The results are shown in Table \ref{table:generation-properties}.
 As \selectro decodes to a bag made up from a predefined set of molecules, those reactants going into the reaction predictor are valid.
 The validity of the final product is not 100\%, as the reaction predictor can make non-valid edits to these molecules, 
 but we see that in a high number of cases the products are valid too.
 Furthermore, what is very encouraging is that the molecules generated often pass the quality filters, giving evidence that the process of building
 molecules up by combining stable reactant building blocks often leads to stable products.

\subsection{Local Optimization}
\label{sect:local-optimization}

As discussed in Section \ref{sect:property-predictor}, when training \selectro we can simultaneously train a property predictor network, mapping from the latent space of \selectro to the QED score of the final product.
 In this section we look at using the gradient information obtainable from this network to do local optimization to find a molecule created from our reactant pool that has a high QED score.

\begin{figure}[h]
\centering

\begin{minipage}{.51\textwidth}
\centering
  \includegraphics[trim={.8em .8em .3em .3em},clip,width=.75\textwidth]{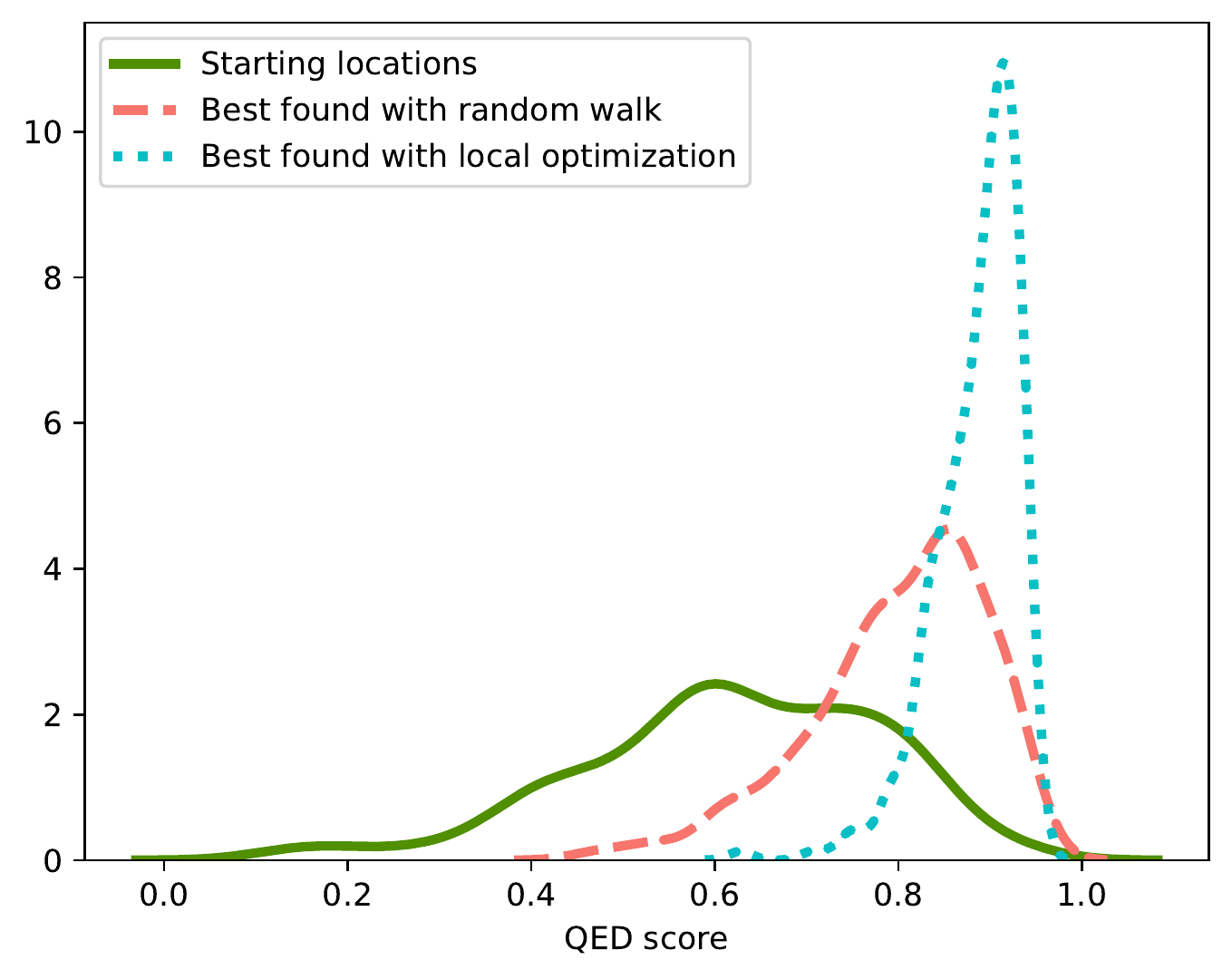}
  \captionof{figure}{KDE plot showing that the distribution of the best QEDs found 
  through local optimization, using our trained property predictor for QEDs,
   has higher mass over higher QED scores compared to the best found from a random walk.
  The starting locations' distribution (sampled from the training data) is shown in green.
  The final products, given a reactant bag are predicted using the MT \citep{schwaller2018molecular}.
}
  \label{fig:local-optimization}
  
\end{minipage}\quad%
\begin{minipage}{.46\textwidth}
\centering
  \includegraphics[trim={.4em .4em .3em .3em},clip,width=.8\textwidth]{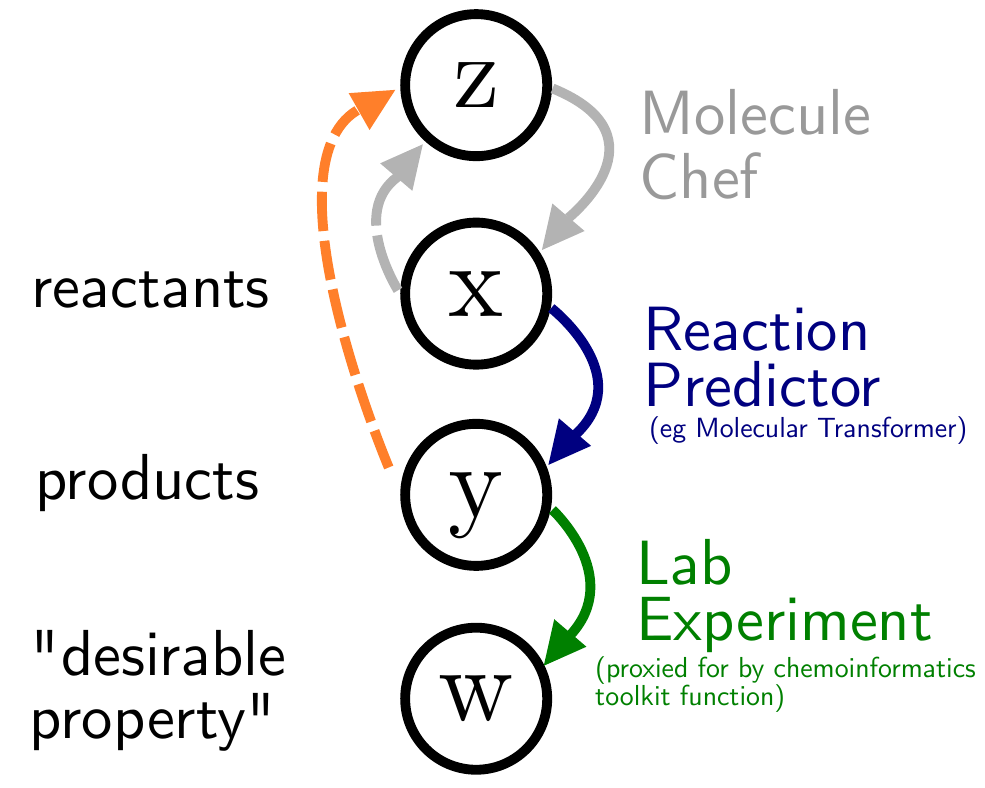}
  \captionof{figure}{
Having learnt a latent space which can map to products through reactants, we can learn a regressor back from the suggested products to the latent space 
(orange dashed \textcolor{orange}{$--$} arrow shown) and couple this with \selectro's decoder to see if we can do retrosynthesis -- the act of computing the reactants that create a particular product.
}
  \label{fig:retro-synthesis-factor-model} 
\end{minipage}%
\end{figure}

We evaluate the local optimization of molecular properties by taking 250 bags of reactants, encoding them into the latent space of \selectro, and then  repeatedly moving in the latent space using the gradient direction of the property predictor until we have decoded ten different reactant bags.
As a comparison we consider instead moving in a random walk until we have also decoded to ten different reaction bags.
In Figure \ref{fig:local-optimization} we look at the distribution of the best QED score found in considering these ten reactant bags, and how this compared to the QEDs started with.

When looking at individual optimization runs, we see that the QEDs vary a lot between different products even if made with similar reactants. 
However, Figure \ref{fig:local-optimization} shows that overall the distribution of the final best found QED scores is improved when purposefully optimizing for this. 
This is encouraging as it gives evidence of the utility of these models for the molecular search problem.

\subsection{Retrosynthesis}

A unique feature of our approach is that we learn a decoder from latent space to a bag of reactants. 
This gives us the ability to do retrosynthesis by training a model to map from products to their associated reactants' representation in latent space and using this in addition to \selectro's decoder to generate a bag of reactants. This process is highlighted in Figure \ref{fig:retro-synthesis-factor-model}.
Although retrosynthesis is a difficult task, with often multiple possible ways to create the same product and with current state-of-the-art approaches built using large reaction databases 
and able to deal with multiple reactions \citep{segler2018planning}, we believe that our model could open up new interesting and exciting approaches to this task. 
We therefore train a small network based on the same graph neural network structure used for \selectro followed by four fully connected layers to regress from products to latent space.

\begin{figure}[h]
\centering
\begin{minipage}{.33\textwidth}

\centering
  \includegraphics[width=0.9\linewidth]{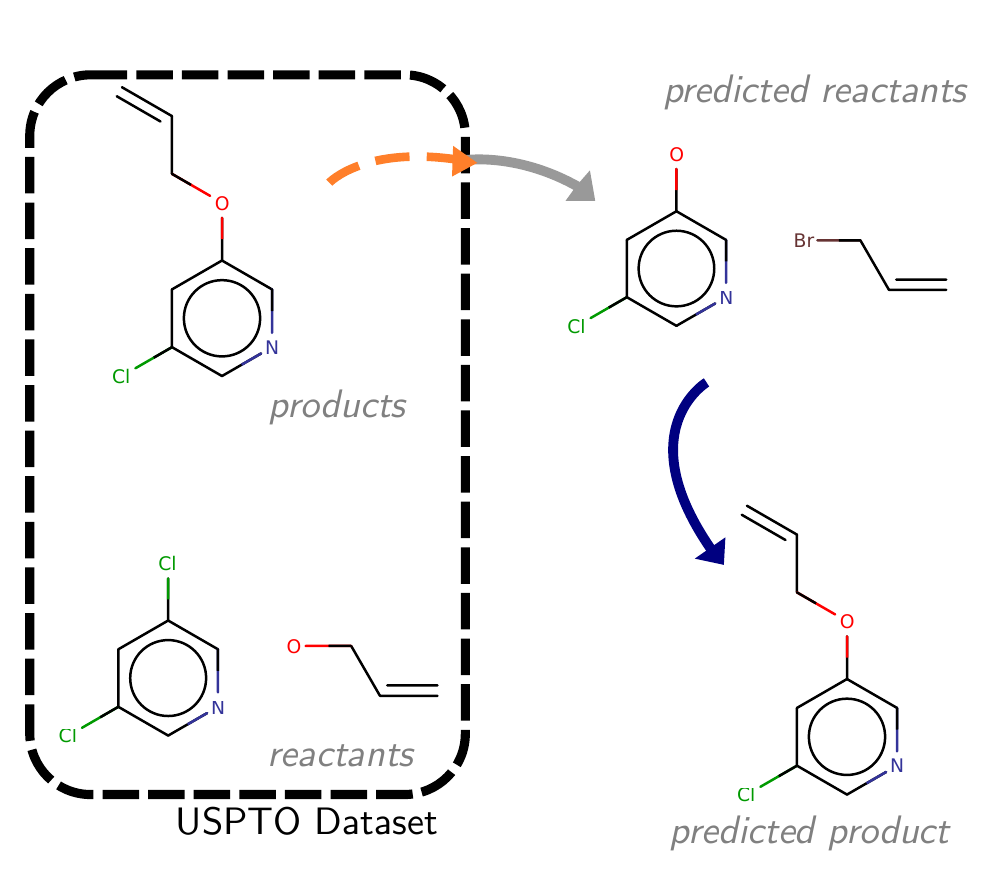}
  \captionof{figure}{
An example of performing retrosynthesis prediction using a trained regressor from products to latent space. 
This reactant-product pair has not been seen in the training set of \selectro. Further examples are shown in the appendix.
}
  \label{fig:retro-eg01}

\end{minipage}\quad%
\begin{minipage}{.6\textwidth}%
  \begin{subfigure}[h]{0.475\textwidth}%
    \includegraphics[width=\textwidth]{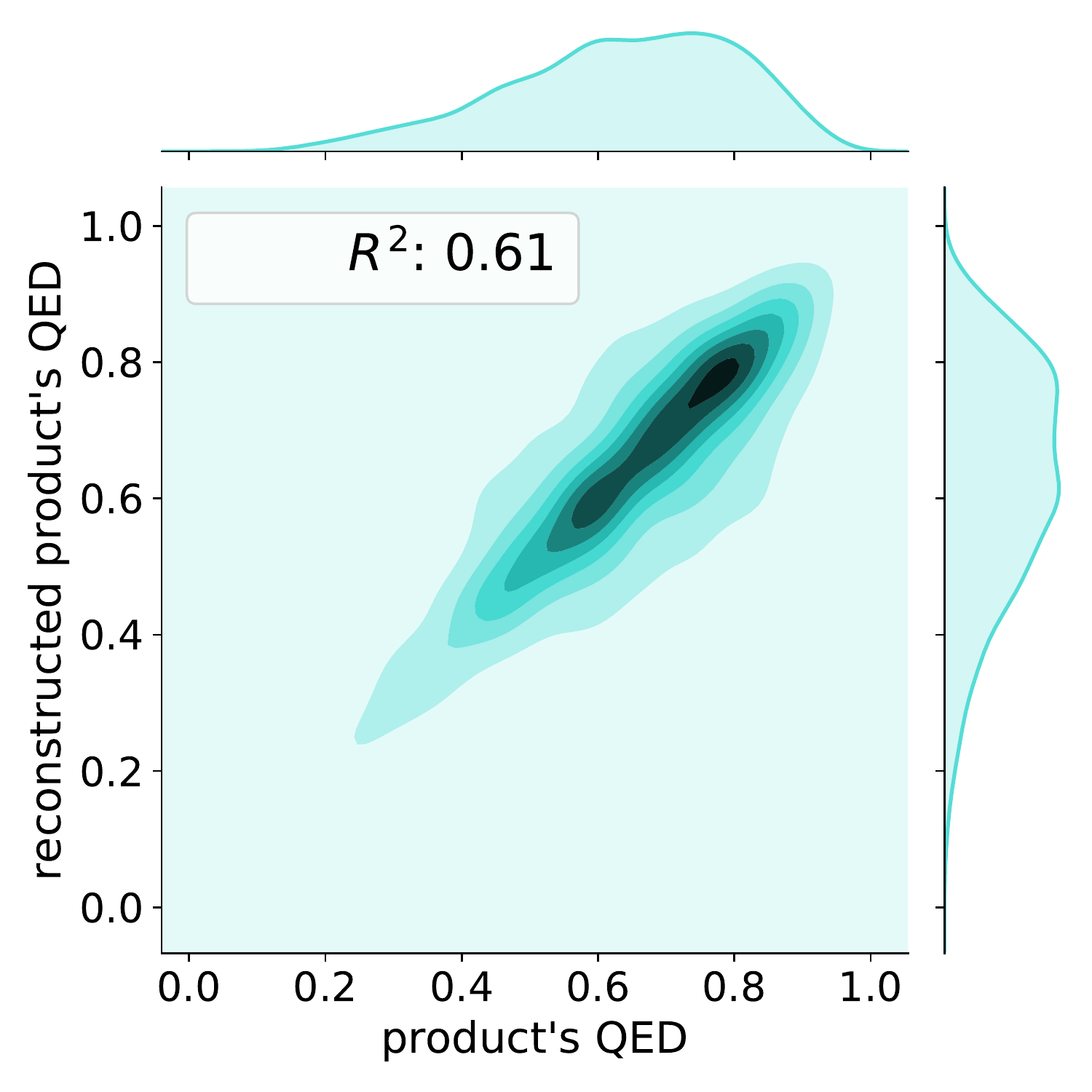}
    \caption{Reachable Products}
    \label{fig:reach}
  \end{subfigure}\hfill
  \begin{subfigure}[h]{0.475\textwidth}
    \includegraphics[width=\textwidth]{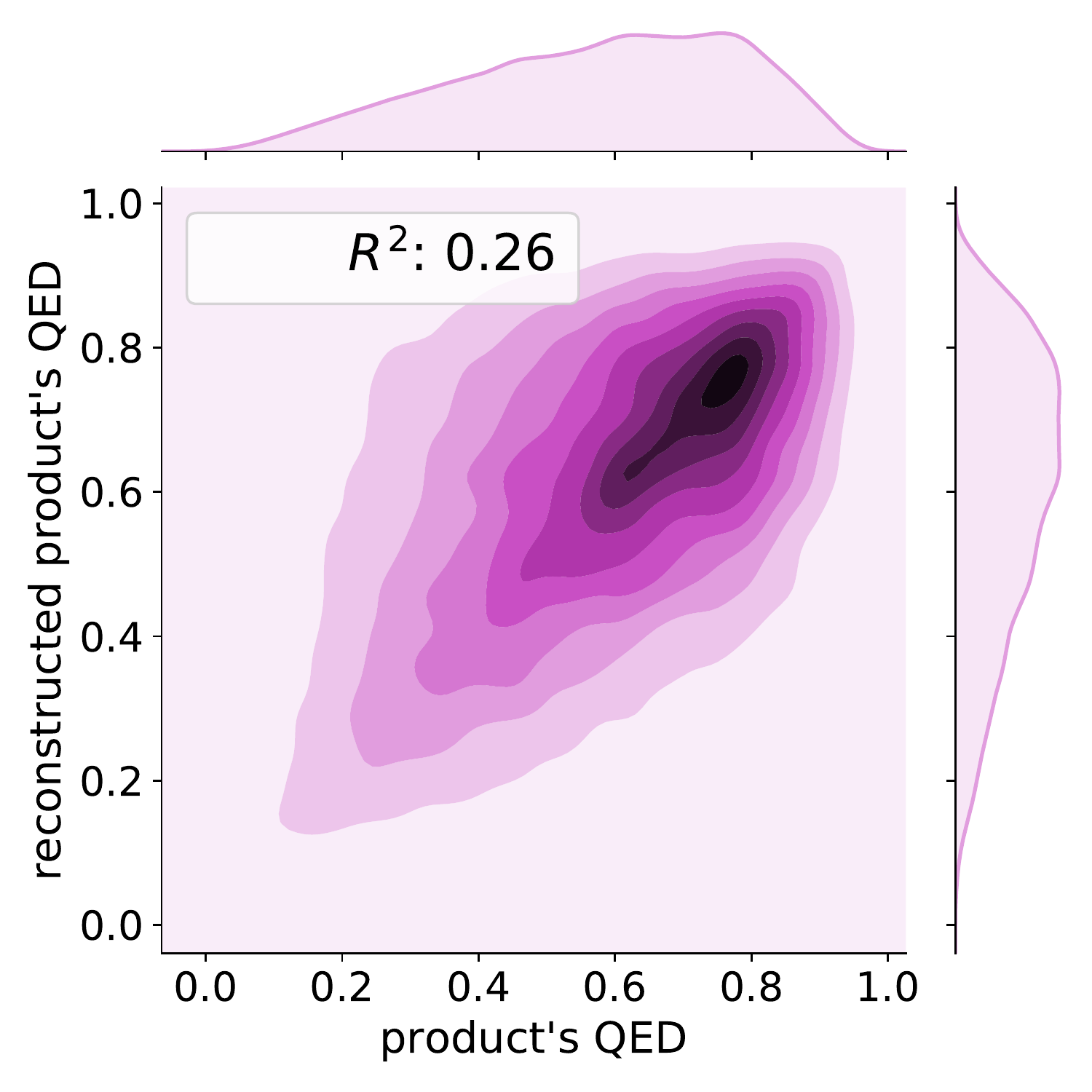}
    \caption{Unreachable Products}
    \label{fig:unreach}
  \end{subfigure}
  \captionof{figure}{Assessing the correlation between the QED scores for the original product and its reconstruction (see text for details).
  We assess on two portions of the test set, products that are made up of only reactants in \selectro's vocabulary are called `Reachable Products',
those that have at least one reactant that is absent are called `Unreachable Products'.} 
  \label{fig:qedReconstructionRetro}
\end{minipage}%
\end{figure}

A few examples of the predicted reactants corresponding to products from reactions in the USPTO test set, but which can be made in one step from the predefined possible reactants, are shown in Figure \ref{fig:retro-eg01} and the appendix.
We see that often this approach, although not always able to suggest the correct whole reactant bag, chooses similar reactants that on reaction produce similar structures to the original product we were trying to synthesize.
While we would not expect this  approach to retrosynthesis to be competitive with complex planning tools, 
we think this provides a promising new approach, 
which could be used to identify bags of reactants that produce molecules similar to a desired target molecule. 
In practice, it would be valuable to be pointed directly to molecules with similar properties to a target molecule if they are easier to make than the target,
since it is the properties of the molecules, and not the actual molecules themselves, that we are after.

With this in mind, we assess our approach in the following way: (1) we take a product and perform retrosynthesis on it to produce a bag of reactants, (2) we transform 
this bag of reactants using the Molecular Transformer to produce a new \emph{reconstructed} product, and then finally (3) we plot the resulting reconstructed product molecule's QED score against the QED score of the initial product. 
We evaluate on a filtered version of \citet{Jin2017-hh}'s test set split of USPTO, where we have filtered out any reactions
which have the exact same reactant and product multisets as a reaction present in the set used to train Molecule Chef.
In addition, we further split this filtered set into two sets:
(i) `Reachable Products', which are reactions in the test set that contain as reactants only molecules that are in \selectro's reactant vocabulary, and (ii) `Unreachable Products', which have at least one reactant molecule
that is not in the vocabulary. 

The results are shown in Figure \ref{fig:qedReconstructionRetro}; overall we see that there is some correlation between the properties
of products and the properties of their reconstructions.
This is more prominent for the reachable products, which we believe is because our latent space is only trained on reachable product reactions and so is better able to model these reactions.
Furthermore, some of the unreachable products may also require reactants that are not available in our pool of easily
available reactants, at least when considering one-step reactions.
However, given that unreachable products
have at least one reactant which is not in Molecule Chef’s vocabulary, we think it is very encouraging that there still is
some, albeit smaller, correlation with the true QED. This is because it shows that our model can suggest molecules with
similar properties made from reactants that are available.

\subsection{Qualitative Quality of Samples}
\label{sect:qualitive_eval}

\begin{figure*}[hbt]
\centering
  \includegraphics[trim={.2em .2em .1em .3em},clip,width=0.87\textwidth]{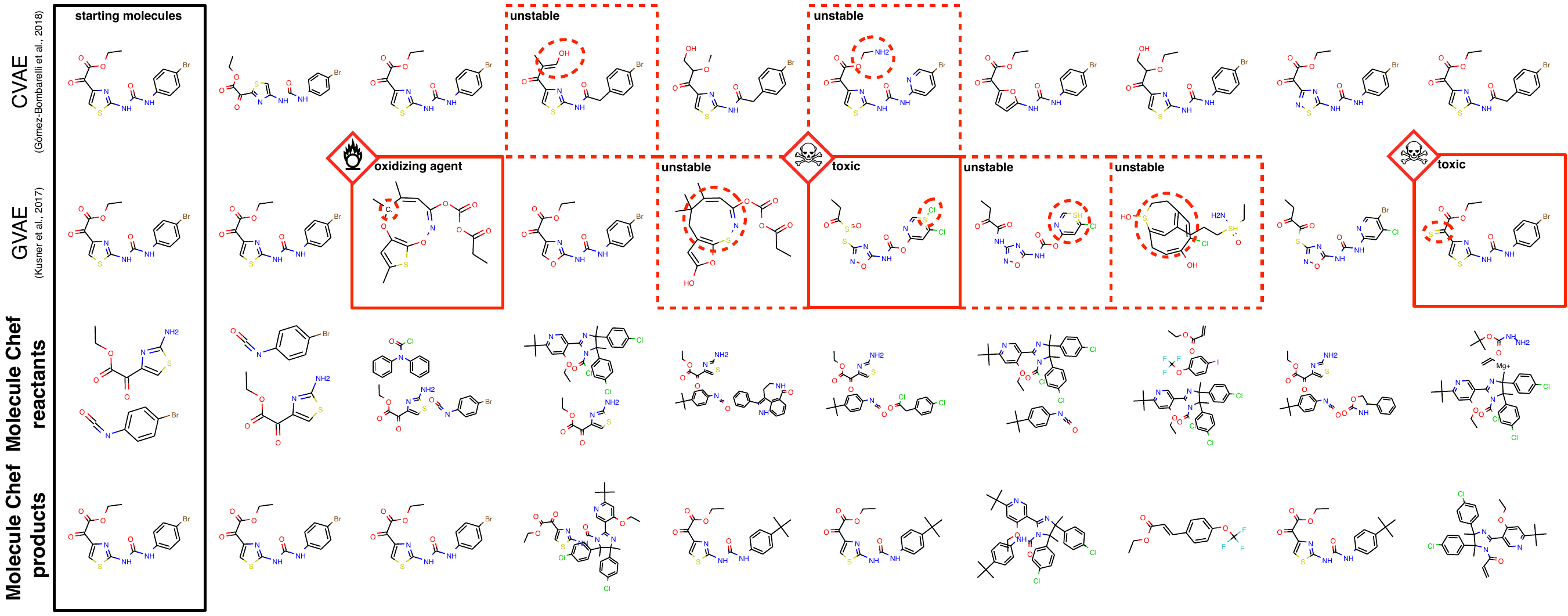}
  \vspace{-2ex}
  \caption{Random walk in latent space. See text for details.}
  \label{fig:evaeDecoder}
\end{figure*}

In Figure~\ref{fig:evaeDecoder} we show molecules generated from a random walk starting from the encoding of a particular molecule (shown in the left-most column). 
We compare the CVAE, GVAE, and \selectro (for \selectro we encode the reactant bag known to generate the same molecule). 
We showed all generated molecules to a domain expert and asked them to evaluate their properties in terms of their \emph{stability}, \emph{toxicity}, \emph{oxidizing power}, \emph{corrosiveness}
(the rationales are provided in more detail in the Appendix). 
Many molecules produced by the CVAE and GVAE show undesirable features, unlike the molecules generated by \selectro.

%% file: tex/conclusion.tex

In this work we have introduced \selectro, a model that generates synthesizable molecules, by considering the products produced as a result of one-step reactions
from a pool of pre-defined reactants. 
By constructing molecules 
through selecting reactants and running chemical reactions, while performing optimization in a continuous latent space, 
we can combine the strengths of previous VAE-based models and classical discrete de-novo design algorithms based on virtual reactions. 
As future work, we hope to explore how to extend our approach to deal with larger reactant vocabularies and multi-step reactions.
This would allow the generation of a wider range of molecules, 
whilst maintaining our approach's advantages of being able to suggest synthetic routes and often producing \emph{semantically valid} molecules.

%% file: tex/appendix.tex

\subsection{Generation Benchmarks on ZINC}

We also ran the baselines for the generation task on the ZINC dataset \citep{irwin2012zinc}. 
The results are shown in Table \ref{table:generation-properties-zinc}.

\begin{table*}[h]
  \caption{
    Table showing generation results for the baseline models when trained on ZINC dataset \citep{irwin2012zinc}.
    The first four result columns show the validity, uniqueness, novelty and normalized quality  (all as \%, higher better)
    of the molecules generated from decoding from 20k random samples from the prior $p(\mathbf z)$.
    Quality is the proportion of molecules that pass the quality filters proposed in \citet[\S3.3]{brown2018guacamol},
    normalized such that the score on the USPTO derived training dataset (used in the main paper) is 100.
     FCD is the Fréchet ChemNet Distance \citep{doi:10.1021/acs.jcim.8b00234}, capturing a notion of distance between
     the generated molecules and the USPTO derived training dataset used in the main paper.
   }
\label{table:generation-properties-zinc}
\begin{center}
  \begin{tabular}{lrrrrr}
    \toprule
    Model Name & Validity & Uniqueness & Novelty & Quality & FCD \\
    \midrule
    AAE \citep{kadurin2017cornucopia, polykovskiy2018molecular}	&	87.64	&	100.00	&	99.99	&	96.12 &	7.27		\\
    CGVAE \citep{Liu2018-ha}	&	100.00	&	95.39	&	96.54		&	43.48 &	15.30 \\
    CVAE \citep{Gomez-Bombarelli2018-ex}	&	0.31	&	40.98	&	24.59	&	128.54 &	40.10	\\
    GVAE \citep{Kusner2017-ry}	&3.66	&	85.23	&	95.08&	38.86 	&	27.31		\\
    LSTM \citep{segler2017generating}	&	95.71	&	99.98	&	99.93	&	108.68 	&	7.99	\\
\bottomrule
  \end{tabular}
    \end{center}
\end{table*}

\FloatBarrier

\subsection{Further Random Walk Examples and Rationales for Expert Annotation}
\paragraph{Rationales for expert labels in Figure 9 in the main text.} Denoted using letters for letters for rows and numbers for columns. A3: Unstable, enol; A5: unstable, aminal; B2: reactive, radical; B4: unstable ring system; B5: toxic, reactive sulfur-chloride bond, unstable ring system; B6: unstable ring system; B7: unstable ring system; B9: toxic: thioketone.

\begin{figure*}[h]
\centering
  \includegraphics[width=0.9\textwidth]{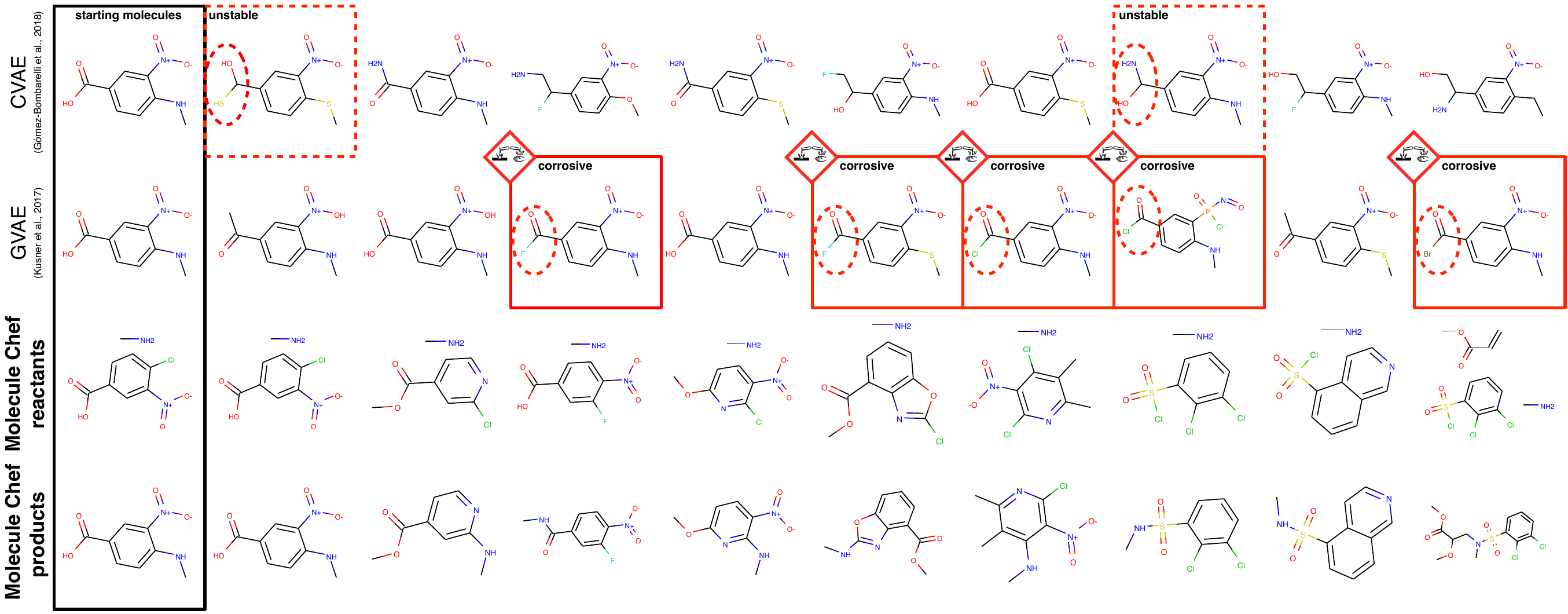}
  \caption{Another example random walk in latent space. See \S 4.4 of main paper
          for further details. The rationals for the labels are (using letters for letters for rows and numbers for columns): A1: Unstable, gemthiolol; A7: unstable, gem-aminohydroxyl; B3: corrosive, acyl fluoride B5: corrosive, acyl fluoride; B6: corrosive, acyl chloride; B7: corrosive, acyl chloride; B9: corrosive, acyl bromide}
  \label{fig:evaeDecoder2}
  \vspace{-2ex}
\end{figure*}


\begin{figure*}[h]
\centering
  \includegraphics[width=0.9\textwidth]{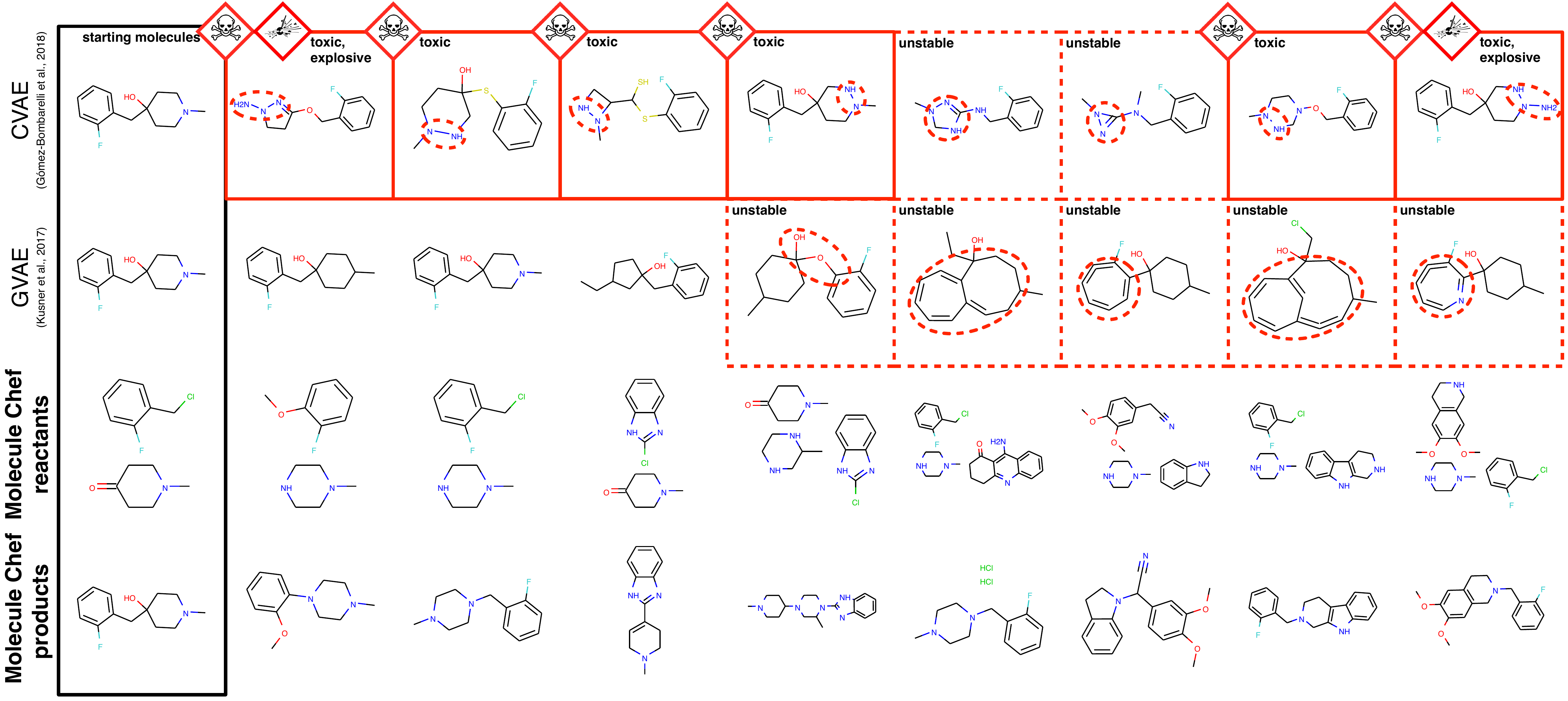}
  \caption{Another example random walk in latent space. See \S 4.4 of main paper
          for further details. The rationals for the labels are (using letters for letters for rows and numbers for columns): A1: toxic, explosive, hydrazine, three consecutive aliphatic nitrogens; A2: toxic, hydrazine; A3: toxic, unstable, hydrazine, hemithioacetal; A4: toxic, hydrazine; A5: unstable, could be oxidized to 1,3,4-triazol, potentially also toxic due to N-N bond/hydrazine; A6: unstable, three-membered ring is antiaromatic; A7: toxic, hydrazine; A8: toxic, explosive, hydrazine, three consecutive aliphatic nitrogens. B4: unstable, hemiacetal; B5-B8: unstable, unfavorable ring systems}
  \label{fig:evaeDecoder2}
  \vspace{-2ex}
\end{figure*}

\FloatBarrier

\subsection{Further Retrosynthesis Results}

In this section we first provide more retrosynthesis examples before also describing an extra experiment in which we try to assess how well the 
retrosynthesis pipeline is at finding molecules with similar properties, even if not reconstructing the correct reactants themselves.

\subsubsection{Further Examples}

\begin{figure*}[h]
    \centering
    \begin{subfigure}[b]{0.41\textwidth}
        \includegraphics[height=5.5cm]{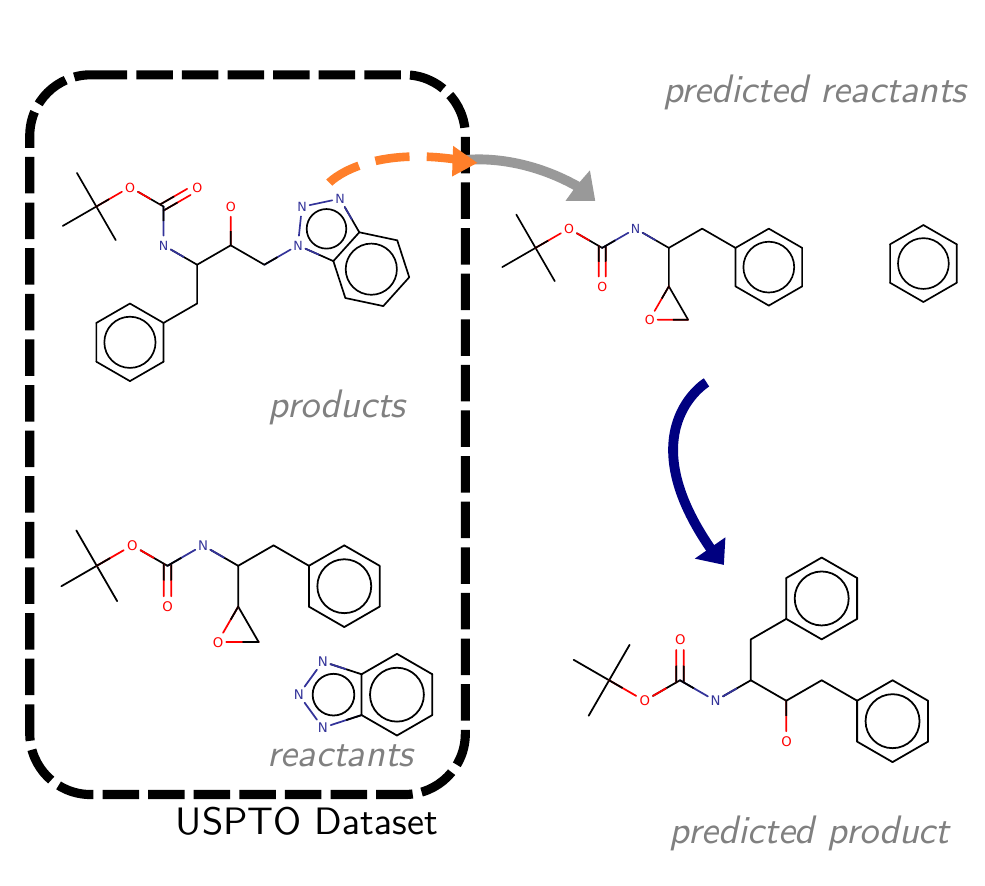}
        \caption{}
    \end{subfigure}
    ~ ~ ~ ~ 
    \begin{subfigure}[b]{0.41\textwidth}
      \includegraphics[height=5.5cm]{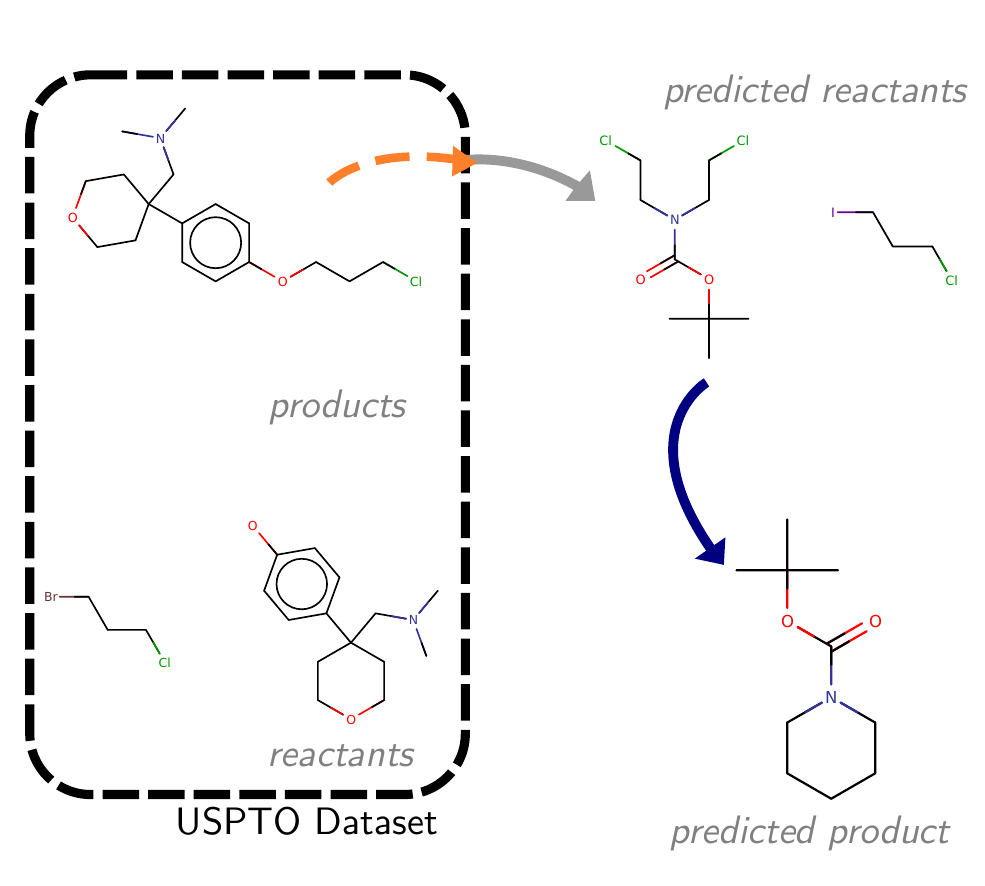}
        \caption{}
    \end{subfigure}

    ~ 
    \begin{subfigure}[b]{0.41\textwidth}
        \includegraphics[height=5.5cm]{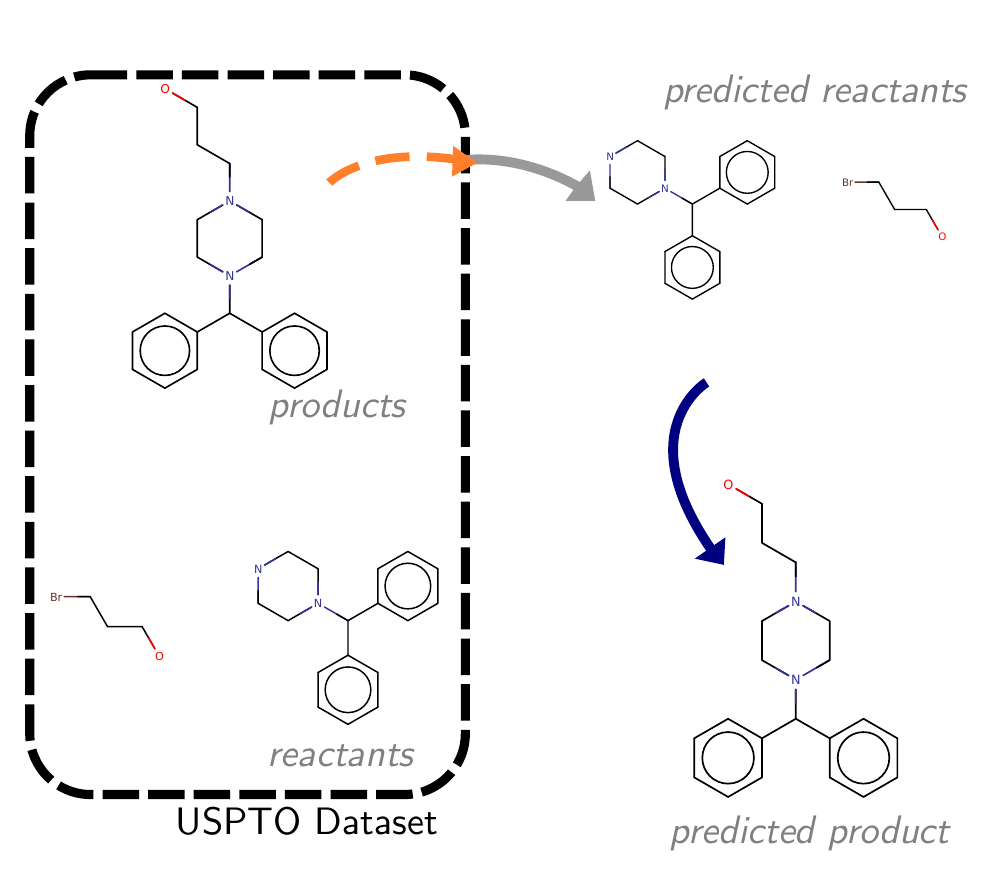}
        \caption{}
    \end{subfigure}
    ~ ~ ~ ~
    \begin{subfigure}[b]{0.41\textwidth}
        \includegraphics[height=5.5cm]{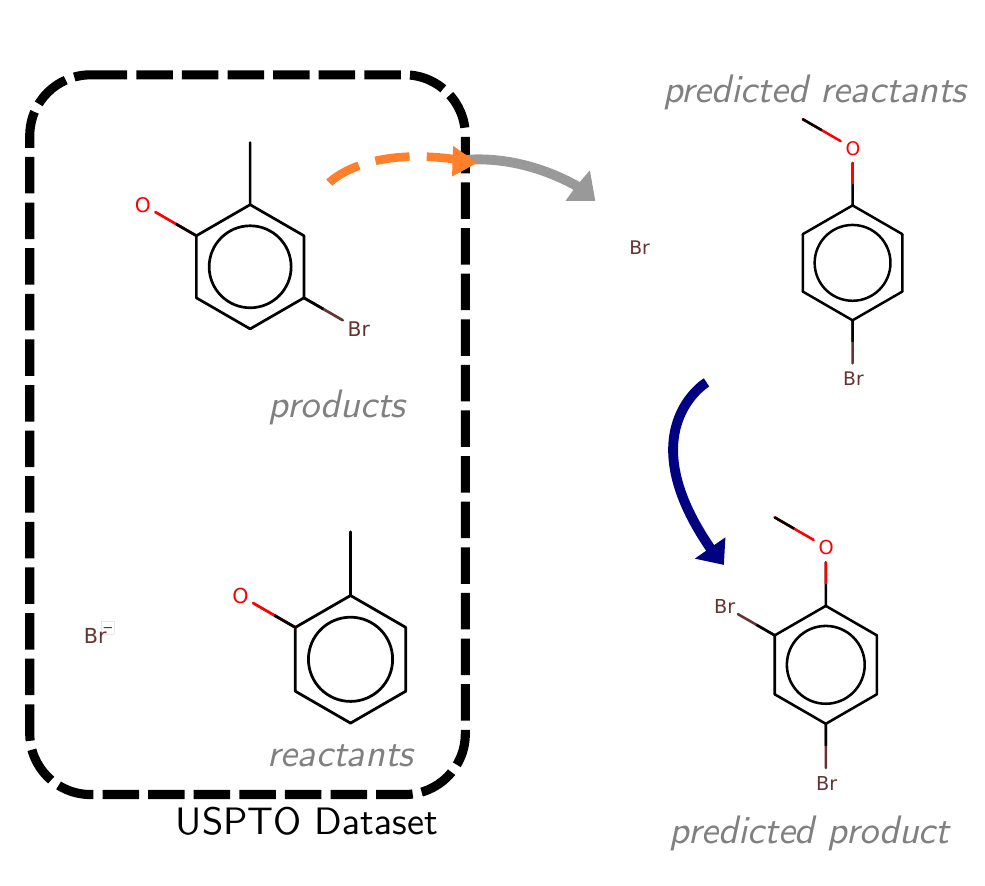}
        \caption{}
    \end{subfigure}

    \caption{Further examples of the predicted reactants associated with a given product for product molecules not in \selectro's training dataset, however with reactants belonging to \selectro's vocabulary (ie Reachable Dataset).
        }\label{fig:more-retro}

\end{figure*}

\begin{figure*}[h]
    \begin{subfigure}[b]{0.41\textwidth}
        \includegraphics[height=5.5cm]{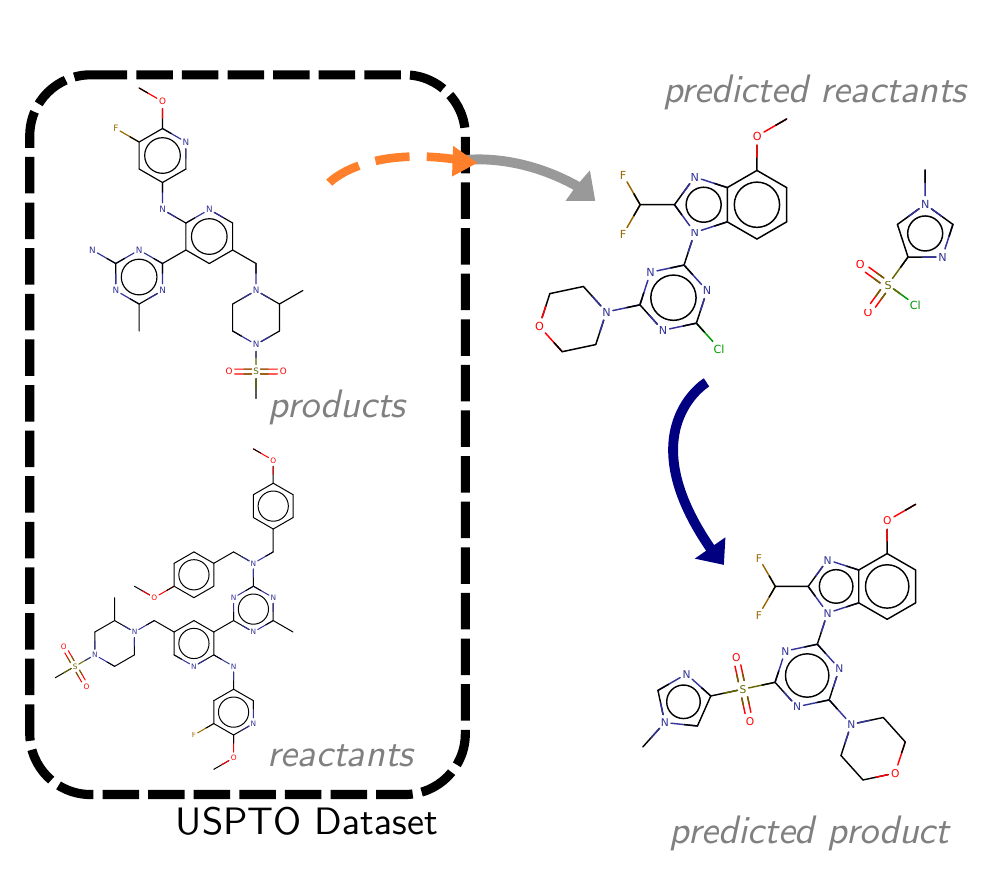}
        \caption{}
    \end{subfigure}
    ~ ~ ~ ~
    \begin{subfigure}[b]{0.41\textwidth}
        \includegraphics[height=5.5cm]{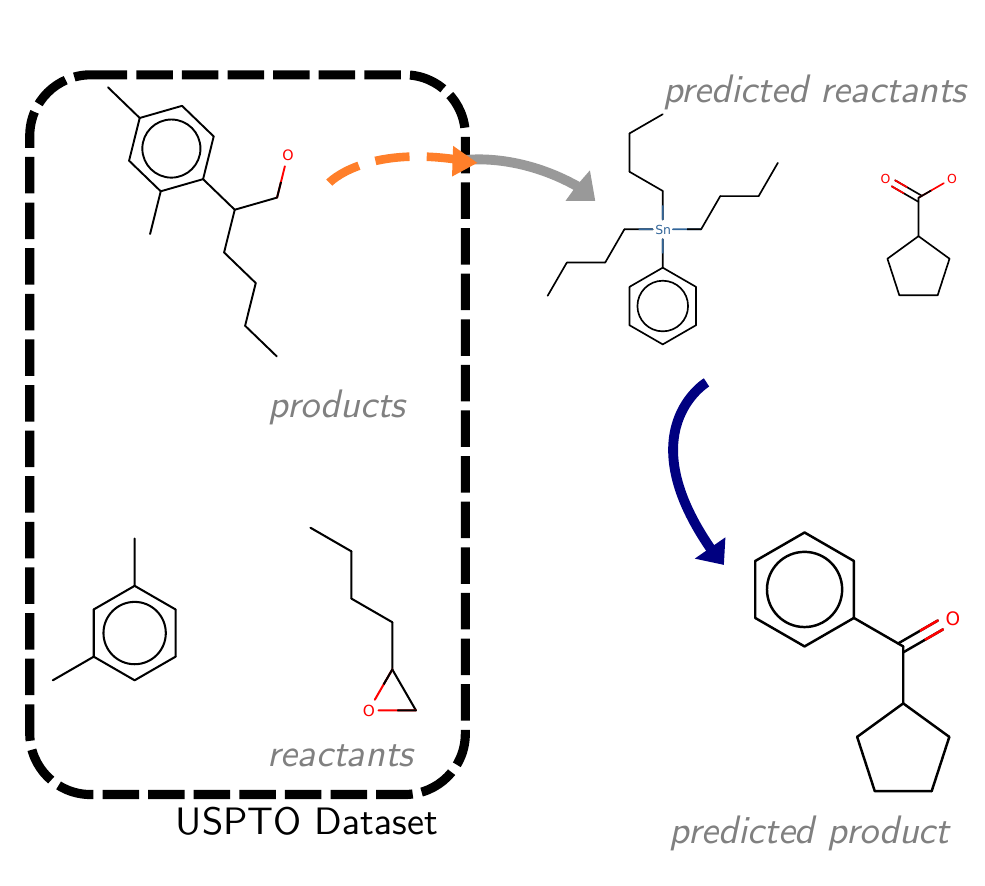}
        \caption{}
    \end{subfigure}

        \caption{Further examples of the predicted reactants associated with a given product for product molecules not in \selectro's training dataset, with at least one reactants not part of \selectro's vocabulary (ie Unreachable Dataset).
        }\label{fig:even-more-retro}

\end{figure*}

\FloatBarrier
\subsubsection{ChemNet Distances between Products and their Reconstructions}

We also consider an experiment for which we analyze the Euclidean distance between the ChemNet embeddings of the product and the reconstructed product
(found by feeding the original product through our retrosynthesis pipeline and then the Molecular Transformer). 
ChemNet embeddings are used when calculating the FCD score \citep{doi:10.1021/acs.jcim.8b00234} between molecule distributions, 
and so hopefully capture various properties of the molecule \citep{C8SC00148K}.
Whilst learning \selectro we include a NN regressor from the latent space to the associated ChemNet embeddings, for which the MSE loss is minimized during training.

To try to establish an idea of how randomly chosen pairs of molecules in our dataset differ from each other, when measured using this metric, we provide a distribution of the distances of random pairs.
This distribution is formed by taking each of the molecules in our dataset (consisting of all the reactants and their associated products) 
and matching it up with another randomly chosen molecule from this set, 
before measuring the Euclidean distance between the embeddings of each of these pairs.

\begin{figure*}[h]
    \begin{subfigure}[t]{0.41\textwidth}
        \includegraphics[width=\textwidth]{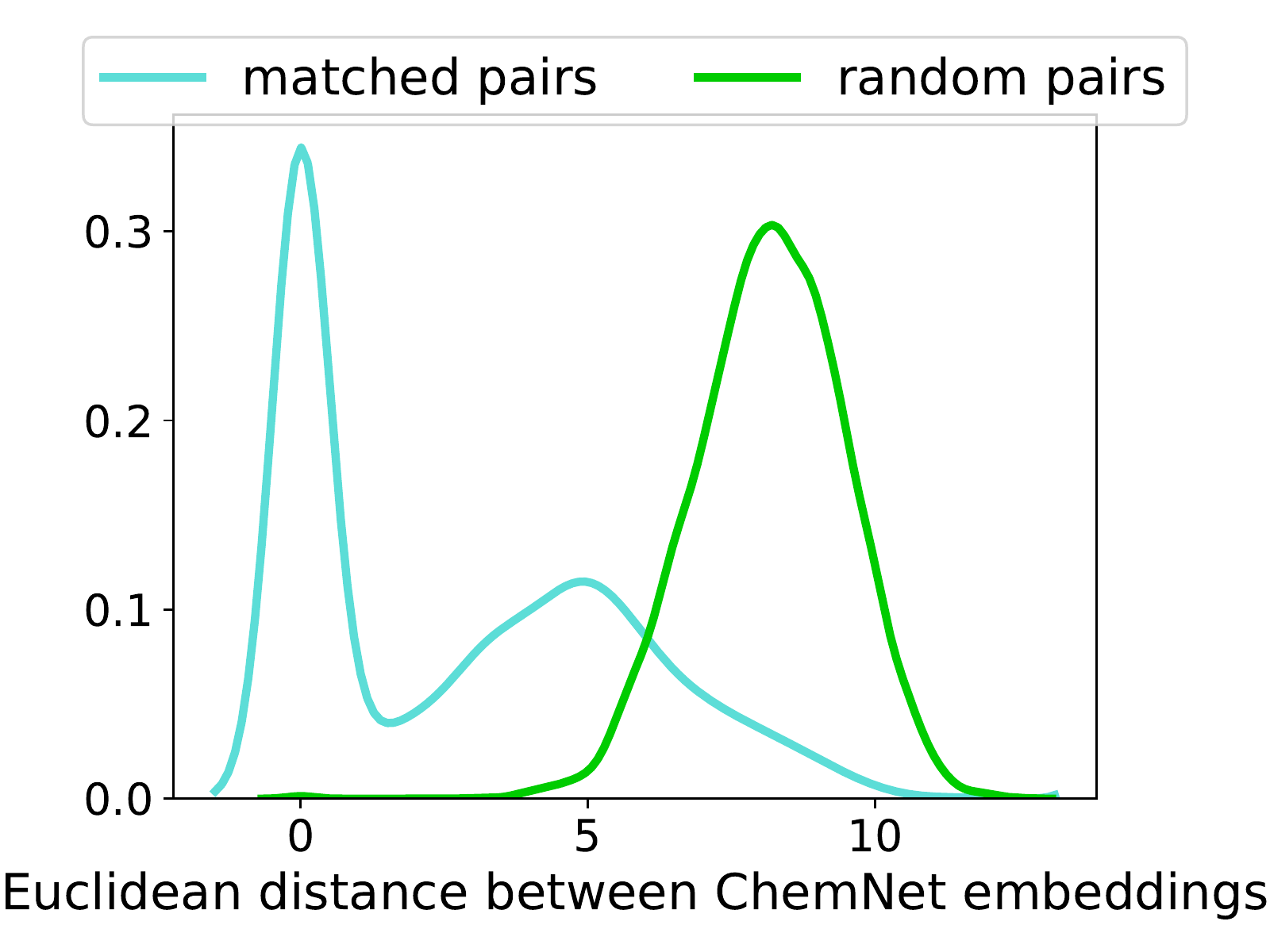}
        \caption{When evaluated on the portion of USPTO test set reactions for which both reactants are present in the \selectro's vocabulary.}
    \end{subfigure}
    ~ ~ ~ ~ ~ ~ ~
    \begin{subfigure}[t]{0.41\textwidth}
        \includegraphics[width=\textwidth]{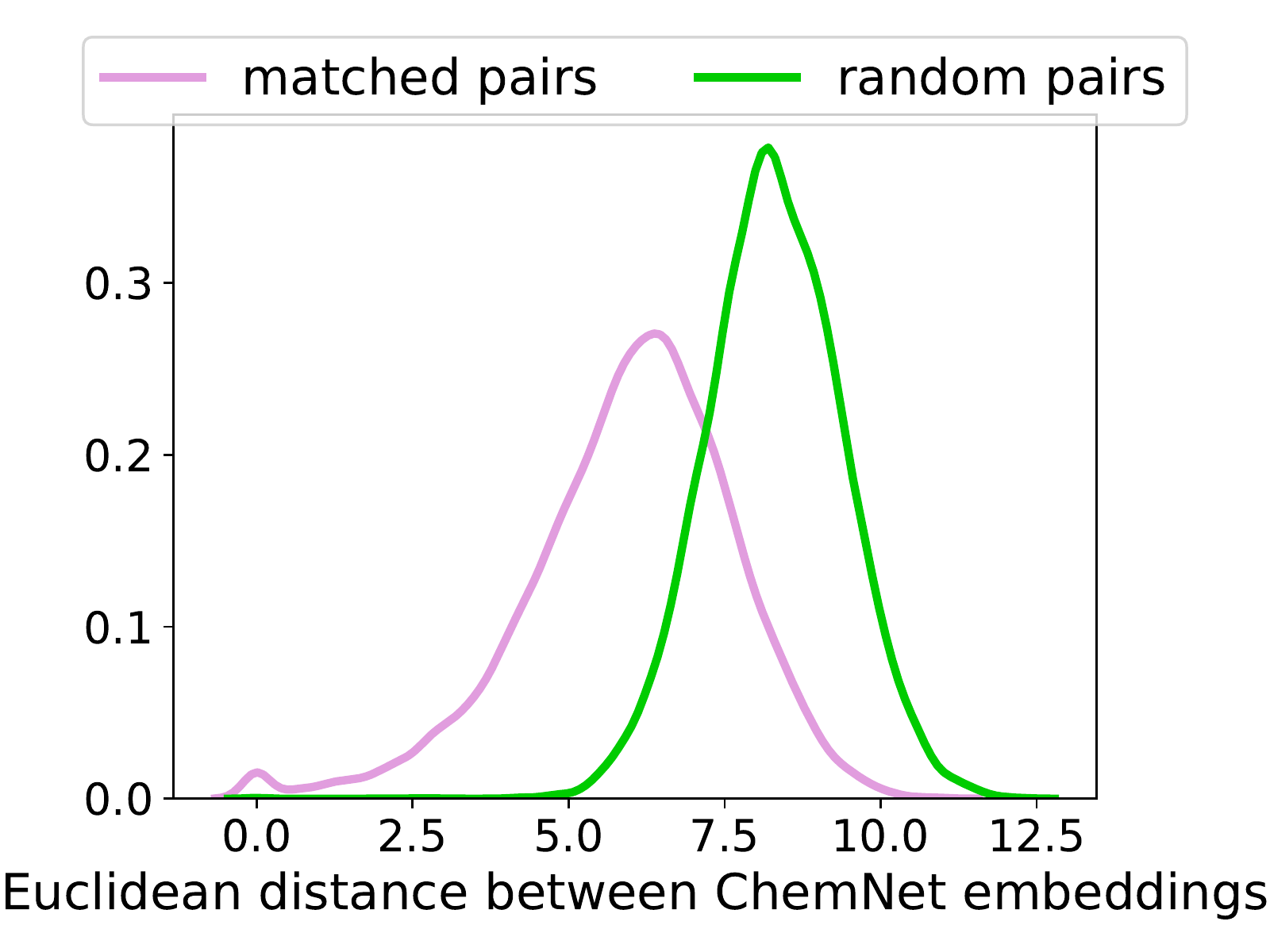}
        \caption{When evaluated on the portion of USPTO test set reactions for which at least one reactant is not present in the \selectro's vocabulary.}
    \end{subfigure}
        \caption{
          KDE plot showing the distribution of the Euclidean distances between the ChemNet embeddings \citep{doi:10.1021/acs.jcim.8b00234} of our product and reconstructed product.
        }\label{fig:chemnet}
\end{figure*}

The results are shown in Figure \ref{fig:chemnet}. We see that the distribution of distances between the products and their reconstructions
has greater mass on smaller distances compared to the random pairs baseline.

\FloatBarrier

\subsection{Details about our Dataset}

In this section we provide further details about the molecules used in training our model and the baselines. 
We also describe details of the molecules used in the retrosynthesis experiments.

For \selectro's vocabulary we use reactants that occur at least 15 times in the USPTO train dataset, as processed and split by \citet{Jin2017-hh}.
This dataset uses reactions collected by \citet{lowe2012extraction} from USPTO patents.
In total we have 4344 reactants, and a training set of 34426 unique reactant bags for which these reactants co-occur. 
Each reactant bag is associated with a product.

For the baselines we train on these reactants and the associated products. This results in a dataset of approximately 37000 unique molecules, containing a wide variety of heavy elements:

 { \tt \{ 'Al', 'B', 'Br', 'C', 'Cl', 'Cr', 'Cu', 'F', 'I', 'K', 'Li', 'Mg', 'Mn', 'N', 'Na', 'O', 'P', 'S', 'Se', 'Si', 'Sn', 'Zn' \} }.

\begin{figure*}[h]
\centering
\includegraphics[width=0.9\textwidth]{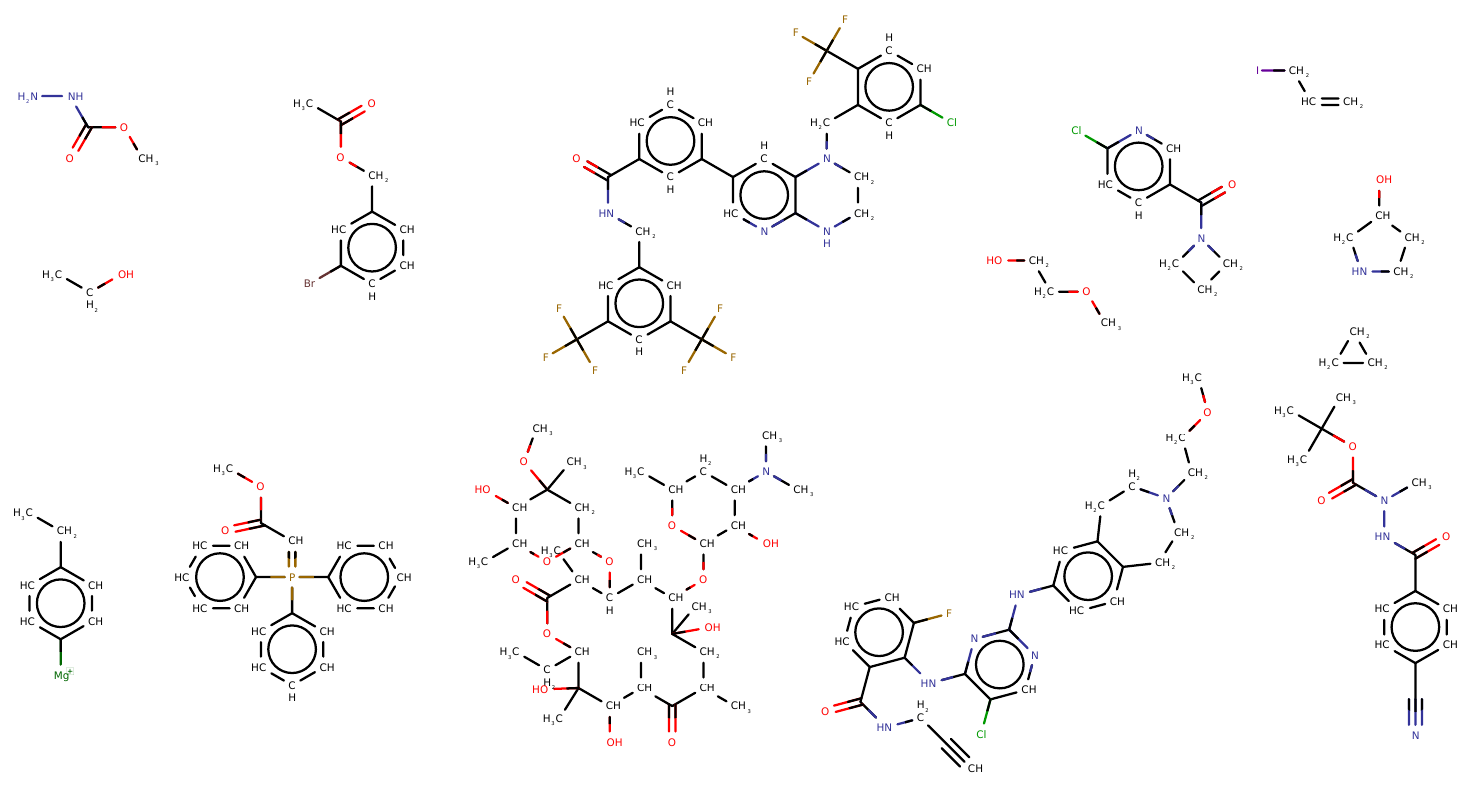}
  \caption{Examples of molecules found in the dataset we use for training the baselines.
    This is a subset of the molecules found in USPTO \citep{lowe2012extraction}.
    It consists of the reactants that the \selectro can produce along with their corresponding products.
    It contains complex molecules with challenging structures to learn.
  }
  \label{fig:uspto_examples}
  \vspace{-2ex}
\end{figure*}

Some examples of the molecules found in the dataset are shown in Figure \ref{fig:uspto_examples}.
Note that the large number of heavy atoms present, as well as the small overall dataset size,
makes a challenging learning task compared to when using some of the more common benchmark datasets used elsewhere (such as ZINC \citep{irwin2012zinc}).

We use examples from the USPTO test dataset when performing the retrosynthesis experiments. 
However, we first filter out any reactions for which the exact same reactant/product multisets tuple is also present in the training data for
\selectro\footnote{After canonicalisation and the removal of reagents, the USPTO train and test dataset has some reactions present in both sets.}.
Then we split the resultant dataset into two subsets.
The first, which we refer to as the reachable dataset, contains only reactants in \selectro's vocabulary.
The second, which we refer to as the unreachable dataset contains reactions with at least one reactant not in the vocabulary.

\FloatBarrier
\subsection{Implementation Details}

\subsection*{Details of \selectro's architecture and parameters}

\begin{figure*}[h]
\centering
\includegraphics[width=\textwidth]{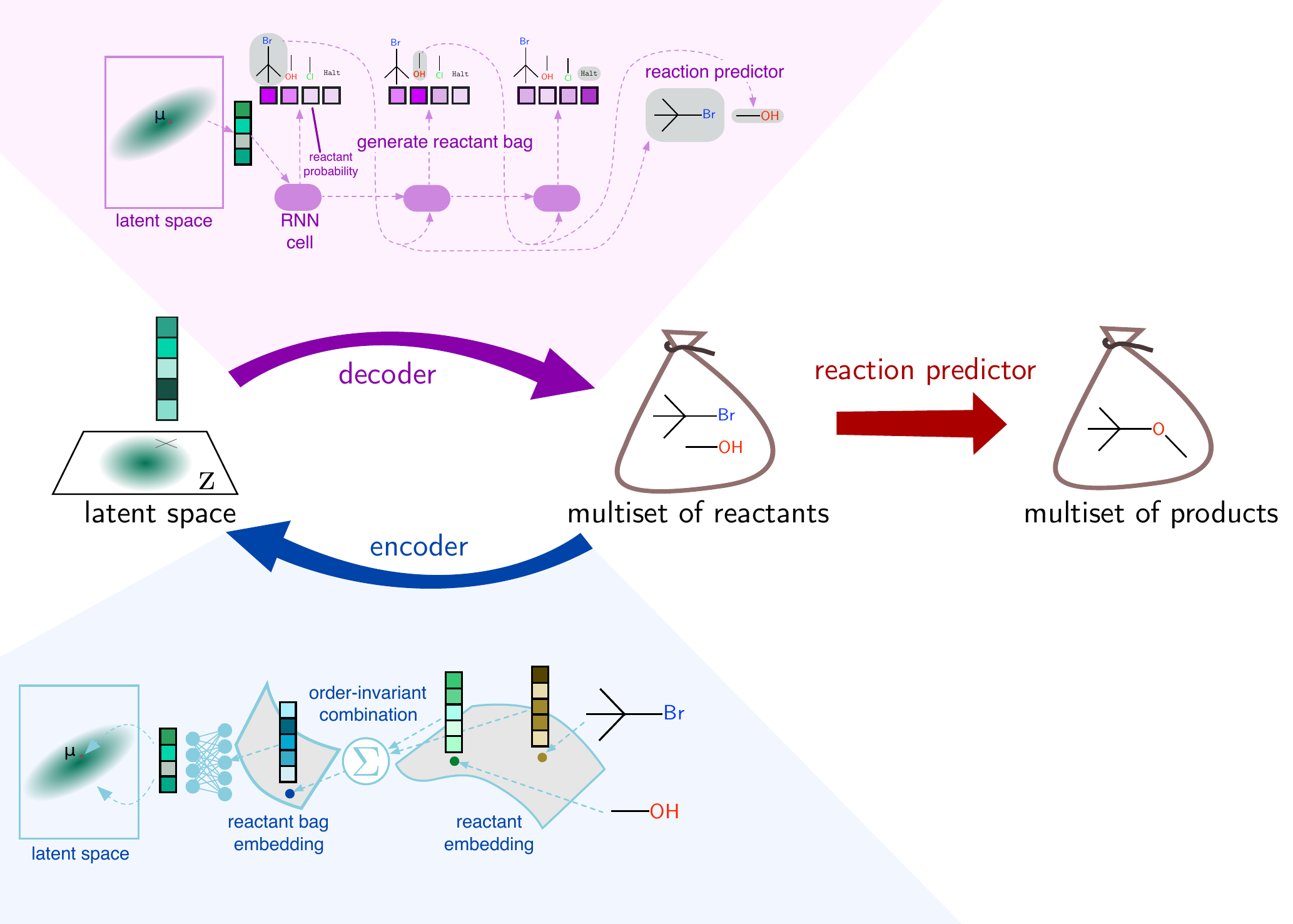}
\caption{Overview of \selectro showing how the encoder and decoder fit together.}
  \label{fig:architecture-overview}
\end{figure*}

An overview of \selectro's architrecture can be seen in Figure \ref{fig:architecture-overview}.
The encoder takes in a multiset of reactants and outputs the parameters of a Gaussian distribution over $\bm{z}$. 
The decoder maps from the latent space to a multiset of reactant molecules. 
Both of these networks rely in turn on vector representations of molecules computed by a graph neural network.
We provide details of these networks' architectures below as well as training details.
Further information can be found in our code available at \url{https://github.com/john-bradshaw/molecule-chef}.

\paragraph{Computing vector representations of molecular graphs using graph neural networks}
For computing the vector representation of molecular graphs we used Gated Graph Neural Networks \citep{li2015gated}, 
with the same network shared in both the encoder and decoder.
We run these networks for 4 propagation steps and the node representations have a dimension of 101. 
We initialise the node reprsentations with the atom features shown in Table \ref{table:atom-features}.
The final step's node representation is projected down to a dimension of 50 by using a learnt linear projection.
Graph level representations are formed from these node representations by performing a weighted sum.

\begin{table}[H]
  \caption{Atom features we use as input to the GGNN. These are calculated using RDKit.}
  \label{table:atom-features}
  \centering
  \begin{tabular}{ll}
    \toprule
    Feature     & Description      \\
    \midrule
    Atom type & 72 possible elements in total, one hot  \\
    Degree     & One hot (0,   1,   2,   3,   4,   5,   6,   7,  10)  \\
    Explicit Valence     & One hot   (0,   1,   2,   3,   4,   5,   6,   7,   8,  10,  12,  14)    \\
    Hybridization & One hot (SP, SP2, SP3, Other) \\
    H count & integer \\
    Electronegativity & float \\
    Atomic number & integer \\
    Part of an aromatic ring & boolean\\
    \bottomrule
  \end{tabular}
\end{table}

\paragraph{Encoder}
The encoder sums the vector representations of the molecules present in the reactant multiset to get a 50 dimensional vector representation of the entire multiset. 
This representation is fed through a single hidden layer NN (with a hidden layer size of 200) to parameterise the mean and diagonal of the covariance matrix of a 25 dimensional multivariate-Gaussian distribution over $\bm{z}$.

\paragraph{Decoder}
The decoder maps from the the latent space, $\bm{z}$, to a multiset of reactants.
It does this through a sequential process, selecting one reactant at a time using a gated recurrent unit (GRU) \citep{cho-etal-2014-learning} RNN.
The parameters used for this GRU are shown in Table \ref{table:decoder-params}.
The initial hidden state of the RNN is set using the result from a learnt linear projection of $\bm{z}$.
The final output of the GRU is fed through a single hidden layer NN (with a hidden size of 128) to form a final output vector.
The dot product of this final output vector is formed with each of the possible reactant embeddings as well as the HALT embedding to form logits for the next output of the decoder.
The embedding of the reactant selected is fed back in as input into the RNN at the next step.

\begin{table}[H]
  \caption{Parameters for GRU used in decoder}
  \label{table:decoder-params}
  \centering
  \begin{tabular}{ll}
    \toprule
    Parameter    & Value      \\
    \midrule
    GRU hidden size & 50 \\
    GRU number of layers & 2 \\
    GRU maximum number of steps & 5 \\
    \bottomrule
  \end{tabular}
\end{table}

\paragraph{Property Predictor} In \S3.2 of the main paper we discuss how we also can train a property predictor from the latent space to a property of interest such as the QED,
while traing the WAE. For the QED property predictor NN we use a fully connected network with two hidden layers, both with dimensionality of 40.
The loss from this network is added to the WAE loss when training the model for
the local optimization and retrosynthesis tasks.

\paragraph{Training} We train the WAE (and property predictor when applicable) for 100 epochs. We use the Adam optimizer \citep{kingma2014adam}, with an initial learning rate of 0.001. 
We decay the learning rate by a factor of 10 every 40 epochs.

\subsubsection*{Implementation Details for the Baselines in Section 4.1 of Main Paper}
For the baselines in the generation section in the main paper we use the following implementations:
\begin{itemize}
  \item CGVAE \citep{Liu2018-ha}: \url{https://github.com/microsoft/constrained-graph-variational-autoencoder}
  \item LSTM \citep{segler2017generating},  : \url{https://github.com/BenevolentAI/guacamol_baselines}
  \item AAE \citep{kadurin2017cornucopia, polykovskiy2018molecular}:  \url{https://github.com/molecularsets/moses/tree/master/moses/aae}
  \item GVAE \citep{Kusner2017-ry}: \url{https://github.com/mkusner/grammarVAE}
  \item CVAE \citep{Gomez-Bombarelli2018-ex}: \url{https://github.com/mkusner/grammarVAE}
\end{itemize}

The LSTM baseline implementation follows \citet{segler2017generating}, which has as its alphabet a list of all individual element symbols, plus special characters used in SMILES strings. 
This differs from the alphabet used by the decoder in the Molecular Transformer \citep{schwaller2018molecular}, 
which instead extracts ``bracketed'' atoms directly from the training set;
this means that a portion of a SMILES string such as \texttt{[OH+]} or \texttt{[NH2-]} would be represented as a single symbol, rather than as a sequence of five symbols.
A regular expression can be used to extract a list of all such sequences from the training data.
Effectively, this makes the trade off of increasing the alphabet size (from 47 to 203 items), while reducing the chance of making syntax errors or suggesting invalid charges.
In practice we found very little qualitative or quantitative difference in the performance of the LSTM model for the two alphabets; 
for sake of consistency with \selectro we report the baseline using the larger alphabet.

For the CGVAE we decide to include element-charge-valence triplets that occur at least 10 times over all the molecules in the training data. 
At generation time we pick one starting node at random.

\subsubsection*{Other Details}

The majority of the experiments for \selectro were run on a NVIDIA Tesla K80 GPU.
For running the Molecular Transformer and CGVAE, we used NVIDIA P100 and P40 GPUs, 
as the latter in particular required a large memory GPU for training on the larger datasets.

For \selectro we have not tried a wide range of hyperparameters.
For the latent dimensionality we initially tried a dimension of 100 before trying and sticking with 25.
Initially, we did not anneal the learning rate but found slightly improved performance by annealing it by a factor of 10 after 40 epochs.
These changes were made after considering the reconstruction error of the model on the validation set (the validation dataset of USPTO restricted to the reactants in \selectro's vocabulary).

%% file: main.bbl
\begin{thebibliography}{59}
\providecommand{\natexlab}[1]{#1}
\providecommand{\url}[1]{\texttt{#1}}
\expandafter\ifx\csname urlstyle\endcsname\relax
  \providecommand{\doi}[1]{doi: #1}\else
  \providecommand{\doi}{doi: \begingroup \urlstyle{rm}\Url}\fi

\bibitem[Alemi et~al.(2018)Alemi, Poole, Fischer, Dillon, Saurous, and
  Murphy]{alemi2018fixing}
Alexander Alemi, Ben Poole, Ian Fischer, Joshua Dillon, Rif~A Saurous, and
  Kevin Murphy.
\newblock Fixing a broken {ELBO}.
\newblock In \emph{International Conference on Machine Learning}, pages
  159--168, 2018.

\bibitem[Assouel et~al.(2018)Assouel, Ahmed, Segler, Saffari, and
  Bengio]{assouel2018defactor}
Rim Assouel, Mohamed Ahmed, Marwin~H Segler, Amir Saffari, and Yoshua Bengio.
\newblock Defactor: Differentiable edge factorization-based probabilistic graph
  generation.
\newblock \emph{arXiv preprint arXiv:1811.09766}, 2018.

\bibitem[Battaglia et~al.(2018)Battaglia, Hamrick, Bapst, Sanchez-Gonzalez,
  Zambaldi, Malinowski, Tacchetti, Raposo, Santoro, Faulkner,
  et~al.]{battaglia2018relational}
Peter~W Battaglia, Jessica~B Hamrick, Victor Bapst, Alvaro Sanchez-Gonzalez,
  Vinicius Zambaldi, Mateusz Malinowski, Andrea Tacchetti, David Raposo, Adam
  Santoro, Ryan Faulkner, et~al.
\newblock Relational inductive biases, deep learning, and graph networks.
\newblock \emph{arXiv preprint arXiv:1806.01261}, 2018.

\bibitem[Bickerton et~al.(2012)Bickerton, Paolini, Besnard, Muresan, and
  Hopkins]{bickerton2012quantifying}
G~Richard Bickerton, Gaia~V Paolini, J{\'e}r{\'e}my Besnard, Sorel Muresan, and
  Andrew~L Hopkins.
\newblock Quantifying the chemical beauty of drugs.
\newblock \emph{Nature chemistry}, 4\penalty0 (2):\penalty0 90, 2012.

\bibitem[Bowman et~al.(2016)Bowman, Vilnis, Vinyals, Dai, Jozefowicz, and
  Bengio]{bowman2015generating}
Samuel~R Bowman, Luke Vilnis, Oriol Vinyals, Andrew~M Dai, Rafal Jozefowicz,
  and Samy Bengio.
\newblock Generating sentences from a continuous space.
\newblock In \emph{Proceedings of The 20th SIGNLL Conference on Computational
  Natural Language Learning}, 2016.

\bibitem[Bradshaw et~al.(2019)Bradshaw, Kusner, Paige, Segler, and
  Hern{\'a}ndez-Lobato]{bradshaw2018generative}
John Bradshaw, Matt~J Kusner, Brooks Paige, Marwin~HS Segler, and
  Jos{\'e}~Miguel Hern{\'a}ndez-Lobato.
\newblock A generative model for electron paths.
\newblock In \emph{{International Conference on Learning Representations}},
  2019.

\bibitem[Brown et~al.(2019)Brown, Fiscato, Segler, and
  Vaucher]{brown2018guacamol}
Nathan Brown, Marco Fiscato, Marwin~H.S. Segler, and Alain~C. Vaucher.
\newblock Guacamol: Benchmarking models for de novo molecular design.
\newblock \emph{Journal of Chemical Information and Modeling}, 59\penalty0
  (3):\penalty0 1096--1108, 2019.
\newblock \doi{10.1021/acs.jcim.8b00839}.

\bibitem[Chevillard and Kolb(2015)]{chevillard2015scubidoo}
Florent Chevillard and Peter Kolb.
\newblock Scubidoo: A large yet screenable and easily searchable database of
  computationally created chemical compounds optimized toward high likelihood
  of synthetic tractability.
\newblock \emph{J. Chem. Inf. Mod.}, 55\penalty0 (9):\penalty0 1824--1835,
  2015.

\bibitem[Cho et~al.(2014)Cho, van Merri{\"e}nboer, Gulcehre, Bahdanau,
  Bougares, Schwenk, and Bengio]{cho-etal-2014-learning}
Kyunghyun Cho, Bart van Merri{\"e}nboer, Caglar Gulcehre, Dzmitry Bahdanau,
  Fethi Bougares, Holger Schwenk, and Yoshua Bengio.
\newblock Learning phrase representations using {RNN} encoder{--}decoder for
  statistical machine translation.
\newblock In \emph{Proceedings of the 2014 Conference on Empirical Methods in
  Natural Language Processing ({EMNLP})}, pages 1724--1734, Doha, Qatar,
  October 2014. Association for Computational Linguistics.
\newblock \doi{10.3115/v1/D14-1179}.
\newblock URL \url{https://www.aclweb.org/anthology/D14-1179}.

\bibitem[Coley et~al.(2019)Coley, Jin, Rogers, Jamison, Jaakkola, Green,
  Barzilay, and Jensen]{coley2019graph}
Connor~W Coley, Wengong Jin, Luke Rogers, Timothy~F Jamison, Tommi~S Jaakkola,
  William~H Green, Regina Barzilay, and Klavs~F Jensen.
\newblock A graph-convolutional neural network model for the prediction of
  chemical reactivity.
\newblock \emph{Chemical Science}, 10\penalty0 (2):\penalty0 370--377, 2019.

\bibitem[Dai et~al.(2018)Dai, Tian, Dai, Skiena, and Song]{dai2018syntax}
Hanjun Dai, Yingtao Tian, Bo~Dai, Steven Skiena, and Le~Song.
\newblock Syntax-directed variational autoencoder for structured data.
\newblock In \emph{International Conference on Learning Representations}, 2018.

\bibitem[De~Cao and Kipf(2018)]{De_Cao2018-sq}
Nicola De~Cao and Thomas Kipf.
\newblock {MolGAN}: An implicit generative model for small molecular graphs.
\newblock In \emph{International Conference on Machine Learning Deep Generative
  Models Workshop}, 2018.

\bibitem[Duvenaud et~al.(2015)Duvenaud, Maclaurin, Iparraguirre, Bombarell,
  Hirzel, Aspuru-Guzik, and Adams]{duvenaud2015convolutional}
David~K Duvenaud, Dougal Maclaurin, Jorge Iparraguirre, Rafael Bombarell,
  Timothy Hirzel, Al{\'a}n Aspuru-Guzik, and Ryan~P Adams.
\newblock Convolutional networks on graphs for learning molecular fingerprints.
\newblock In \emph{Advances in neural information processing systems}, pages
  2224--2232, 2015.

\bibitem[Gardner et~al.(2014)Gardner, Kusner, Xu, Weinberger, and
  Cunningham]{gardner2014bayesian}
Jacob~R Gardner, Matt~J Kusner, Zhixiang~Eddie Xu, Kilian~Q Weinberger, and
  John~P Cunningham.
\newblock Bayesian optimization with inequality constraints.
\newblock In \emph{International Conference on Machine Learning}, pages
  937--945, 2014.

\bibitem[G{\'o}mez-Bombarelli et~al.(2018)G{\'o}mez-Bombarelli, Wei, Duvenaud,
  Hern{\'a}ndez-Lobato, S{\'a}nchez-Lengeling, Sheberla, Aguilera-Iparraguirre,
  Hirzel, Adams, and Aspuru-Guzik]{Gomez-Bombarelli2018-ex}
Rafael G{\'o}mez-Bombarelli, Jennifer~N Wei, David Duvenaud, Jos{\'e}~Miguel
  Hern{\'a}ndez-Lobato, Benjam{\'\i}n S{\'a}nchez-Lengeling, Dennis Sheberla,
  Jorge Aguilera-Iparraguirre, Timothy~D Hirzel, Ryan~P Adams, and Al{\'a}n
  Aspuru-Guzik.
\newblock Automatic chemical design using a {Data-Driven} continuous
  representation of molecules.
\newblock \emph{ACS Cent Sci}, 4\penalty0 (2):\penalty0 268--276, February
  2018.

\bibitem[Guimaraes et~al.(2017)Guimaraes, Sanchez-Lengeling, Outeiral, Farias,
  and Aspuru-Guzik]{guimaraes2017objective}
Gabriel~Lima Guimaraes, Benjamin Sanchez-Lengeling, Carlos Outeiral, Pedro
  Luis~Cunha Farias, and Al{\'a}n Aspuru-Guzik.
\newblock Objective-reinforced generative adversarial networks ({ORGAN}) for
  sequence generation models.
\newblock \emph{arXiv preprint arXiv:1705.10843}, 2017.

\bibitem[Hartenfeller and Schneider(2011)]{hartenfeller2011enabling}
Markus Hartenfeller and Gisbert Schneider.
\newblock Enabling future drug discovery by de novo design.
\newblock \emph{Wiley Interdisc. Rev. Comp. Mol. Sci.}, 1\penalty0
  (5):\penalty0 742--759, 2011.

\bibitem[Hu et~al.(2011)Hu, Peng, Kostrowicki, and Kuki]{hu2011leap}
Qiyue Hu, Zhengwei Peng, Jaroslav Kostrowicki, and Atsuo Kuki.
\newblock Leap into the {Pfizer} global virtual library ({PGVL}) space:
  creation of readily synthesizable design ideas automatically.
\newblock In \emph{Chemical Library Design}, pages 253--276. Springer, 2011.

\bibitem[Irwin et~al.(2012)Irwin, Sterling, Mysinger, Bolstad, and
  Coleman]{irwin2012zinc}
John~J Irwin, Teague Sterling, Michael~M Mysinger, Erin~S Bolstad, and Ryan~G
  Coleman.
\newblock {ZINC}: a free tool to discover chemistry for biology.
\newblock \emph{Journal of chemical information and modeling}, 52\penalty0
  (7):\penalty0 1757--1768, 2012.

\bibitem[Janz et~al.(2018)Janz, van~der Westhuizen, Paige, Kusner, and
  Hern{\'a}ndez-Lobato]{janz2017learning}
David Janz, Jos van~der Westhuizen, Brooks Paige, Matt~J Kusner, and
  Jos{\'e}~Miguel Hern{\'a}ndez-Lobato.
\newblock Learning a generative model for validity in complex discrete
  structures.
\newblock In \emph{International Conference on Learning Representations}, 2018.

\bibitem[Jin et~al.(2017)Jin, Coley, Barzilay, and Jaakkola]{Jin2017-hh}
Wengong Jin, Connor~W Coley, Regina Barzilay, and Tommi Jaakkola.
\newblock Predicting organic reaction outcomes with {Weisfeiler-Lehman}
  network.
\newblock In \emph{{Advances in Neural Information Processing Systems}}, 2017.

\bibitem[Jin et~al.(2018)Jin, Barzilay, and Jaakkola]{Jin2018-aa}
Wengong Jin, Regina Barzilay, and Tommi Jaakkola.
\newblock Junction tree variational autoencoder for molecular graph generation.
\newblock In \emph{International Conference on Machine Learning}, 2018.

\bibitem[Jin et~al.(2019)Jin, Yang, Barzilay, and Jaakkola]{jin2018learning}
Wengong Jin, Kevin Yang, Regina Barzilay, and Tommi Jaakkola.
\newblock Learning multimodal graph-to-graph translation for molecular
  optimization.
\newblock In \emph{International Conference on Learning Representations}, 2019.

\bibitem[Johnson(2017)]{Johnson2017-pd}
Daniel~D Johnson.
\newblock Learning graphical state transitions.
\newblock In \emph{{International Conference on Learning Representations}},
  2017.

\bibitem[Kadurin et~al.(2017)Kadurin, Aliper, Kazennov, Mamoshina, Vanhaelen,
  Khrabrov, and Zhavoronkov]{kadurin2017cornucopia}
Artur Kadurin, Alexander Aliper, Andrey Kazennov, Polina Mamoshina, Quentin
  Vanhaelen, Kuzma Khrabrov, and Alex Zhavoronkov.
\newblock The cornucopia of meaningful leads: Applying deep adversarial
  autoencoders for new molecule development in oncology.
\newblock \emph{Oncotarget}, 8\penalty0 (7):\penalty0 10883, 2017.

\bibitem[Kajino(2019)]{pmlr-v97-kajino19a}
Hiroshi Kajino.
\newblock Molecular hypergraph grammar with its application to molecular
  optimization.
\newblock In \emph{International Conference on Machine Learning}, pages
  3183--3191, 2019.

\bibitem[Kayala et~al.(2011)Kayala, Azencott, Chen, and
  Baldi]{kayala2011learning}
Matthew~A Kayala, Chlo{\'e}-Agathe Azencott, Jonathan~H Chen, and Pierre Baldi.
\newblock Learning to predict chemical reactions.
\newblock \emph{Journal of chemical information and modeling}, 51\penalty0
  (9):\penalty0 2209--2222, 2011.

\bibitem[Kingma and Ba(2015)]{kingma2014adam}
Diederik Kingma and Jimmy Ba.
\newblock Adam: A method for stochastic optimization.
\newblock In \emph{{International Conference on Learning Representations}},
  2015.

\bibitem[Kingma and Welling(2014)]{kingma2013auto}
Diederik~P Kingma and Max Welling.
\newblock Auto-encoding variational {Bayes}.
\newblock In \emph{International Conference on Learning Representations}, 2014.

\bibitem[Kusner et~al.(2017)Kusner, Paige, and
  Hern{\'a}ndez-Lobato]{Kusner2017-ry}
Matt~J Kusner, Brooks Paige, and Jos{\'e}~Miguel Hern{\'a}ndez-Lobato.
\newblock Grammar variational autoencoder.
\newblock In \emph{International Conference on Machine Learning}, 2017.

\bibitem[Li et~al.(2016)Li, Tarlow, Brockschmidt, and Zemel]{li2015gated}
Yujia Li, Daniel Tarlow, Marc Brockschmidt, and Richard Zemel.
\newblock Gated graph sequence neural networks.
\newblock \emph{International Conference on Learning Representations}, 2016.

\bibitem[Li et~al.(2018)Li, Vinyals, Dyer, Pascanu, and Battaglia]{Li2018-zg}
Yujia Li, Oriol Vinyals, Chris Dyer, Razvan Pascanu, and Peter Battaglia.
\newblock Learning deep generative models of graphs.
\newblock \emph{arXiv preprint arXiv:1803.03324}, March 2018.

\bibitem[Liu et~al.(2018)Liu, Allamanis, Brockschmidt, and Gaunt]{Liu2018-ha}
Qi~Liu, Miltiadis Allamanis, Marc Brockschmidt, and Alexander~L Gaunt.
\newblock Constrained graph variational autoencoders for molecule design.
\newblock In \emph{Advances in neural information processing systems}, 2018.

\bibitem[Lowe(2012)]{lowe2012extraction}
Daniel~Mark Lowe.
\newblock \emph{Extraction of chemical structures and reactions from the
  literature}.
\newblock PhD thesis, University of Cambridge, 2012.

\bibitem[Mayr et~al.(2018)Mayr, Klambauer, Unterthiner, Steijaert, Wegner,
  Ceulemans, Clevert, and Hochreiter]{C8SC00148K}
Andreas Mayr, Günter Klambauer, Thomas Unterthiner, Marvin Steijaert, Jörg~K.
  Wegner, Hugo Ceulemans, Djork-Arné Clevert, and Sepp Hochreiter.
\newblock Large-scale comparison of machine learning methods for drug target
  prediction on chembl.
\newblock \emph{Chem. Sci.}, 9:\penalty0 5441--5451, 2018.
\newblock \doi{10.1039/C8SC00148K}.
\newblock URL \url{http://dx.doi.org/10.1039/C8SC00148K}.

\bibitem[Merkwirth and Lengauer(2005)]{merkwirth2005automatic}
Christian Merkwirth and Thomas Lengauer.
\newblock Automatic generation of complementary descriptors with molecular
  graph networks.
\newblock \emph{Journal of chemical information and modeling}, 45\penalty0
  (5):\penalty0 1159--1168, 2005.

\bibitem[Morgan(1965)]{morgan1965generation}
HL~Morgan.
\newblock The generation of a unique machine description for chemical
  structures-a technique developed at chemical abstracts service.
\newblock \emph{Journal of Chemical Documentation}, 5\penalty0 (2):\penalty0
  107--113, 1965.

\bibitem[Nicolaou et~al.(2016)Nicolaou, Watson, Hu, and
  Wang]{nicolaou2016proximal}
Christos~A Nicolaou, Ian~A Watson, Hong Hu, and Jibo Wang.
\newblock The proximal lilly collection: Mapping, exploring and exploiting
  feasible chemical space.
\newblock \emph{Journal of chemical information and modeling}, 56\penalty0
  (7):\penalty0 1253--1266, 2016.

\bibitem[Polykovskiy et~al.(2018)Polykovskiy, Zhebrak, Sanchez-Lengeling,
  Golovanov, Tatanov, Belyaev, Kurbanov, Artamonov, Aladinskiy, Veselov,
  Kadurin, Nikolenko, Aspuru-Guzik, and Zhavoronkov]{polykovskiy2018molecular}
Daniil Polykovskiy, Alexander Zhebrak, Benjamin Sanchez-Lengeling, Sergey
  Golovanov, Oktai Tatanov, Stanislav Belyaev, Rauf Kurbanov, Aleksey
  Artamonov, Vladimir Aladinskiy, Mark Veselov, Artur Kadurin, Sergey
  Nikolenko, Alan Aspuru-Guzik, and Alex Zhavoronkov.
\newblock {M}olecular {S}ets ({MOSES}): {A} {B}enchmarking {P}latform for
  {M}olecular {G}eneration {M}odels.
\newblock \emph{arXiv preprint arXiv:1811.12823}, 2018.

\bibitem[Preuer et~al.(2018)Preuer, Renz, Unterthiner, Hochreiter, and
  Klambauer]{doi:10.1021/acs.jcim.8b00234}
Kristina Preuer, Philipp Renz, Thomas Unterthiner, Sepp Hochreiter, and
  G{\"u}nter Klambauer.
\newblock Fr{\'e}chet chemnet distance: A metric for generative models for
  molecules in drug discovery.
\newblock \emph{Journal of Chemical Information and Modeling}, 58\penalty0
  (9):\penalty0 1736--1741, 2018.
\newblock \doi{10.1021/acs.jcim.8b00234}.

\bibitem[Pyzer-Knapp et~al.(2015)Pyzer-Knapp, Suh, G{\'o}mez-Bombarelli,
  Aguilera-Iparraguirre, and Aspuru-Guzik]{pyzer2015high}
Edward~O Pyzer-Knapp, Changwon Suh, Rafael G{\'o}mez-Bombarelli, Jorge
  Aguilera-Iparraguirre, and Al{\'a}n Aspuru-Guzik.
\newblock What is high-throughput virtual screening? {A} perspective from
  organic materials discovery.
\newblock \emph{Annual Review of Materials Research}, 45:\penalty0 195--216,
  2015.

\bibitem[Rarey and Stahl(2001)]{rarey2001similarity}
Matthias Rarey and Martin Stahl.
\newblock Similarity searching in large combinatorial chemistry spaces.
\newblock \emph{Journal of Computer-Aided Molecular Design}, 15\penalty0
  (6):\penalty0 497--520, 2001.

\bibitem[RDKit, online()]{rdkit}
RDKit, online.
\newblock {RDK}it: Open-source cheminformatics.
\newblock \url{http://www.rdkit.org}.
\newblock [Online; accessed 01-February-2018].

\bibitem[Rezende et~al.(2014)Rezende, Mohamed, and
  Wierstra]{rezende2014stochastic}
Danilo~Jimenez Rezende, Shakir Mohamed, and Daan Wierstra.
\newblock Stochastic backpropagation and approximate inference in deep
  generative models.
\newblock In \emph{International Conference on Machine Learning}, pages
  1278--1286, 2014.

\bibitem[Samanta et~al.(2019)Samanta, Abir, Jana, Chattaraj, Ganguly, and
  Rodriguez]{samanta2019nevae}
Bidisha Samanta, DE~Abir, Gourhari Jana, Pratim~Kumar Chattaraj, Niloy Ganguly,
  and Manuel~Gomez Rodriguez.
\newblock {NeVAE}: A deep generative model for molecular graphs.
\newblock In \emph{Proceedings of the AAAI Conference on Artificial
  Intelligence}, volume~33, pages 1110--1117, 2019.

\bibitem[Schneider and Schneider(2016)]{schneider2016novo}
Petra Schneider and Gisbert Schneider.
\newblock De novo design at the edge of chaos: Miniperspective.
\newblock \emph{J. Med. Chem.}, 59\penalty0 (9):\penalty0 4077--4086, 2016.

\bibitem[Schwaller et~al.(2018)Schwaller, Gaudin, L{\'a}nyi, Bekas, and
  Laino]{schwaller2017found}
Philippe Schwaller, Th{\'e}ophile Gaudin, D{\'a}vid L{\'a}nyi, Costas Bekas,
  and Teodoro Laino.
\newblock {``Found in Translation''}: predicting outcomes of complex organic
  chemistry reactions using neural sequence-to-sequence models.
\newblock \emph{Chem. Sci.}, 9:\penalty0 6091--6098, 2018.
\newblock \doi{10.1039/C8SC02339E}.

\bibitem[Schwaller et~al.(2019)Schwaller, Laino, Gaudin, Bolgar, Hunter, Bekas,
  and Lee]{schwaller2018molecular}
Philippe Schwaller, Teodoro Laino, Théophile Gaudin, Peter Bolgar,
  Christopher~A. Hunter, Costas Bekas, and Alpha~A. Lee.
\newblock Molecular transformer: A model for uncertainty-calibrated chemical
  reaction prediction.
\newblock \emph{ACS Central Science}, 5\penalty0 (9):\penalty0 1572--1583,
  2019.
\newblock \doi{10.1021/acscentsci.9b00576}.

\bibitem[Segler and Waller(2017)]{segler2017neural}
Marwin~HS Segler and Mark~P Waller.
\newblock Neural-symbolic machine learning for retrosynthesis and reaction
  prediction.
\newblock \emph{Chemistry--A European Journal}, 23\penalty0 (25):\penalty0
  5966--5971, 2017.

\bibitem[Segler et~al.(2017)Segler, Kogej, Tyrchan, and
  Waller]{segler2017generating}
Marwin~HS Segler, Thierry Kogej, Christian Tyrchan, and Mark~P Waller.
\newblock Generating focused molecule libraries for drug discovery with
  recurrent neural networks.
\newblock \emph{ACS Cent. Sci.}, 4\penalty0 (1):\penalty0 120--131, 2017.

\bibitem[Segler et~al.(2018)Segler, Preuss, and Waller]{segler2018planning}
Marwin~HS Segler, Mike Preuss, and Mark~P Waller.
\newblock Planning chemical syntheses with deep neural networks and symbolic
  {AI}.
\newblock \emph{Nature}, 555\penalty0 (7698):\penalty0 604, 2018.

\bibitem[Shoichet(2004)]{shoichet2004virtual}
Brian~K Shoichet.
\newblock Virtual screening of chemical libraries.
\newblock \emph{Nature}, 432\penalty0 (7019):\penalty0 862, 2004.

\bibitem[Simonovsky and Komodakis(2018)]{Simonovsky2018-md}
Martin Simonovsky and Nikos Komodakis.
\newblock {GraphVAE}: Towards generation of small graphs using variational
  autoencoders.
\newblock In V{\v{e}}ra K{\r{u}}rkov{\'a}, Yannis Manolopoulos, Barbara Hammer,
  Lazaros Iliadis, and Ilias Maglogiannis, editors, \emph{Artificial Neural
  Networks and Machine Learning -- ICANN 2018}, pages 412--422, Cham, 2018.
  Springer International Publishing.
\newblock ISBN 978-3-030-01418-6.

\bibitem[Snoek et~al.(2012)Snoek, Larochelle, and Adams]{snoek2012practical}
Jasper Snoek, Hugo Larochelle, and Ryan~P Adams.
\newblock Practical {Bayesian} optimization of machine learning algorithms.
\newblock In \emph{Advances in neural information processing systems}, pages
  2951--2959, 2012.

\bibitem[Tolstikhin et~al.(2018)Tolstikhin, Bousquet, Gelly, and
  Schoelkopf]{tolstikhin2017wasserstein}
Ilya Tolstikhin, Olivier Bousquet, Sylvain Gelly, and Bernhard Schoelkopf.
\newblock Wasserstein auto-encoders.
\newblock In \emph{International Conference on Learning Representations}, 2018.

\bibitem[van Hilten et~al.(2019)van Hilten, Chevillard, and
  Kolb]{van2019virtual}
Niek van Hilten, Florent Chevillard, and Peter Kolb.
\newblock Virtual compound libraries in computer-assisted drug discovery.
\newblock \emph{Journal of chemical information and modeling}, 2019.

\bibitem[Wei et~al.(2016)Wei, Duvenaud, and Aspuru-Guzik]{wei2016neural}
Jennifer~N Wei, David Duvenaud, and Al{\'a}n Aspuru-Guzik.
\newblock Neural networks for the prediction of organic chemistry reactions.
\newblock \emph{ACS central science}, 2\penalty0 (10):\penalty0 725--732, 2016.

\bibitem[Weininger(1988)]{weininger1988smiles}
David Weininger.
\newblock {SMILES}, a chemical language and information system. 1.
  {Introduction} to methodology and encoding rules.
\newblock \emph{Journal of chemical information and computer sciences},
  28\penalty0 (1):\penalty0 31--36, 1988.

\bibitem[You et~al.(2018)You, Liu, Ying, Pande, and Leskovec]{you2018graph}
Jiaxuan You, Bowen Liu, Rex Ying, Vijay Pande, and Jure Leskovec.
\newblock Graph convolutional policy network for goal-directed molecular graph
  generation.
\newblock In \emph{Advances in Neural Information Processing Systems}, 2018.

\end{thebibliography}
